\documentclass[14pt, a4paper]{article}
\usepackage[a4paper, left=3cm, right=3cm, top=3cm, bottom=2cm]{geometry}

\usepackage{amssymb}
\usepackage{amsmath}
\usepackage{txfonts}
\usepackage{mathdots}
 \usepackage{graphicx}
\usepackage{caption}
\usepackage{subcaption}
\usepackage{lineno,hyperref}
\usepackage{changepage}
\usepackage{setspace}
\usepackage{algorithm,algpseudocode}
\usepackage{enumerate}
\usepackage{cite}
\usepackage{tabularx,booktabs,caption,paralist}
\usepackage[table]{xcolor}
\usepackage{multirow,paralist}

\DeclareGraphicsExtensions{.pdf,.png,.jpg}

\begin{document}

\title{Energy-Aware JPEG Image Compression: A Multi-Objective Approach}

\author{Seyed Jalaleddin Mousavirad$^1$ and Luís A. Alexandre$^2$  \\
\ \\
$^1$Universidade da Beira Interior, Covilhã, Portugal\\
$^2$NOVA LINCS, Universidade da Beira Interior, Covilhã, Portugal}

\date{}

\maketitle

\begin {abstract}
Customer satisfaction is crucially affected by energy consumption in mobile devices. One of the most energy-consuming parts of an application is images. While different images with different quality consume different amounts of energy, there are no straightforward methods to calculate the energy consumption of an operation in a typical image. This paper, first, investigates that there is a correlation between energy consumption and image quality as well as image file size. Therefore, these two can be considered as a proxy for energy consumption. Then, we propose a multi-objective strategy to enhance image quality and reduce image file size based on the quantisation tables in JPEG image compression. To this end, we have used two general multi-objective metaheuristic approaches: scalarisation and Pareto-based. Scalarisation methods find a single optimal solution based on combining different objectives, while Pareto-based techniques aim to achieve a set of solutions. In this paper, we embed our strategy into five scalarisation algorithms, including energy-aware multi-objective genetic algorithm (EnMOGA), energy-aware multi-objective particle swarm optimisation (EnMOPSO), energy-aware multi-objective differential evolution (EnMODE), energy-aware multi-objective evolutionary strategy (EnMOES), and energy-aware multi-objective pattern search (EnMOPS). Also, two Pareto-based methods, including a non-dominated sorting genetic algorithm (NSGA-II) and a reference-point-based NSGA-II (NSGA-III) are used for the embedding scheme, and two Pareto-based algorithms, EnNSGAII and EnNSGAIII, are presented. Experimental studies show that the performance of the baseline algorithm is improved by embedding the proposed strategy into metaheuristic algorithms. In particular, EnMOGA, EnMOPS, and EnNSGA-II can perform competitively, among others. Furthermore, we statistically verify the proposed algorithm's effectiveness based on the Wilcoxon-signed rank test. Finally, a sensitivity analysis of the parameters is provided. The source code for reproducing the results is available in: \url{https://github.com/SeyedJalaleddinMousavirad/MultiobjectiveJPEGImageCompression}.

\end {abstract}

\section{Introduction}
\label{sec:Intro}
Mobile devices such as smartphones and tablets are ubiquitous and receiving much attention for their energy efficiency since customer satisfaction relies heavily on battery uptime. In addition, battery uptime plays a crucial role for developers since anomalous draining usually warrants negative app store ratings~\cite{APP_01}.

Several studies~\cite{Energy_APP_01,Energy_APP_05,Energy_APP_09,Energy_APP_10} have concentrated on documenting energy-aware programming trends in the context of Android, the leading mobile ecosystem, and finding better alternatives. But images have not been seriously discussed, while they are one of the most important components of mobile software, particularly in games. 

JPEG (\textbf{J}oint \textbf{P}hotographic \textbf{E}xperts \textbf{G}roup) format is the most commonly used method of compression for digital images, and is based on the Discrete Cosine Transform(DCT)~\cite{DCT_original}. The process of JPEG compression is started with representation of the original uncompressed image in
YCbCr colour space, where Y, Cb and Cr indicate luminance, blue and red chrominance components, respectively; and each component is handled independently. We shall simply use the luminance component, Y, for the sake of simplicity, while for other components, the process is the same. The image's component Y is divided into $8 \times 8$ blocks, each of which is separately modified. Before using the DCT, the $8 \times 8$ blocks are zero-shifted by deducting 128 from the element values. Then, each modified block is quantised. The primary mechanism for compression, quantisation, also results in information loss due to the representation of the DCT coefficients. Each block may be effectively entropy encoded after quantisation~\cite{JPEG01}, with no information being lost in the process.

The quantisation table (QT) plays a crucial role in the JPEG image compression. Annex K variant~\cite{Annex_Jpeg}, the most important variant of JPEG implementation, employs two quantisation tables, called luminance quantisation table (LQT) and chrominance quantisation table (CQT). The main responsibility of these two is to quantise the DCT coefficient blocks of luminance and chrominance elements, respectively. The process of finding proper values for both quantisation tables is a challenging and difficult task since each image needs its own table, although most implementations use a conventional value for the tables. 

To design the best quantisation table, meta-heuristic algorithms (MA) such as genetic algorithm (GA)~\cite{GA_Main_Ref} and particle swarm optimisation(PSO)~\cite{PSO_Main_Paper} can be used. MAs are iterative, stochastic, and problem-independent algorithms that solve an optimisation problem by using a number of operators to guide the search process. Also, they can provide a close to optimal solution, but they are not able to guarantee a global optimum solution.
 


In one of the first efforts to use MA for the construction of the JPEG QTs, \cite{JPEG_GA} proposed a GA algorithm to find the quantisation table so that the chromosome is an array of size 64, while the objective function is the mean square error between the original image and the compressed image. In another work, \cite{JPEG_GA2} incorporated GA to design a JPEG image quantisation table to compress iris images in iris recognition systems. \cite{JPEG_GA3} proposed a knowledge-based GA to find the quantisation table. To this end, image characteristics and knowledge about image compression are integrated into the GA algorithm. Differential evolution (DE) is used to design the quantisation table in another paper~\cite{JPEG_DE01}, and they showed that DE could outperform canonical GA. Another study~\cite{JPEG_DE02} proposes a knowledge-based DE to improve the performance of DE. From the literature, we can also observe some other MA algorithms for designing QTs in JPEG, such as simulated annealing (SA)~\cite{JPEG_PSO_SA,JPEG_SA01}, DE algorithm~\cite{JPEG_DE03}, particle swarm optimisation (PSO)~\cite{JPEG_PSO_SA}, firework algorithm~\cite{JPEG_firework}, and firefly algorithm~\cite{JPEG_FA01}. Also, some researchers try to combine different MAs to improve the performance of designing QTs. For instance, \cite{JPEG_FA_TLBO} proposed a combination of FA and teaching-learning-based optimisation to select the QT. Also, reducing time complexity is considered in a few papers. For example, \cite{Time_Jpeg} employs a surrogate model-based DE algorithm to reduce computation time for optimising the QT.    

While images are one of the primary sources of energy consumption in smartphones, it is challenging to measure the amount of energy consumed for a specific operation in a typical image. Most of the current methods in the literature can measure the power of a battery or, at best, for a particular application~\cite{Petra}. To tackle this, we used the energy profiler of Android Studio and Plot Digitiser software, manually and not in an automatic way, to verify that image quality and file size play a crucial role in the energy consumption of an application. In other words, smaller file sizes and lower image quality consume less energy. A developer has two main goals in selecting an image: 1) they tend to select an image with high quality, and 2) they prefer to choose an image with smaller file size. As a result, there are two conflicting criteria for a mobile developer. Since an operation's energy for an image cannot be calculated as a straightforward process, file size and quality can act as a proxy for energy consumption.

These two criteria, file size and image quality, are two conflicting objectives. Therefore, multi-objective metaheuristic optimisation (MOMO) algorithms can be used to tackle this issue. MOMO addresses optimising a problem based on two or more conflicting objective. There are two general approaches for solving a multi-objective problem, namely, scalarisation and Pareto-based approaches~\cite{Survey_MOO}. Scalarisation approaches solve a multi-objective problem by converting it into a single-objective problem, while Pareto-based approaches generate a set of optimal solutions. 

This paper proposes an energy-aware JPEG Image compression strategy. The main characteristics of this paper are as follows:
\begin{enumerate}
	\item We investigate, based on an energy profiler, that there is a high correlation between energy consumption and image quality. Such a condition is also valid for image file size.
	\item We propose a multi-objective strategy for handling both image file size and image quality.
	\item The multi-objective strategy is embedded into five scalarisation methods, including genetic algorithm (GA), differential evolution (DE), particle swarm optimisation (PSO), evolutionary strategy (ES), and pattern search (PS). Therefore, five scalarisation-based multi-objective techniques for JPEG image compression are introduced, namely, EnMOGA, EnMODE, EnMOPSO, EnMOES, and EnMOPS.
	\item We also embed the proposed strategy into two well-known Pareto-based approaches, the non-dominated sorting genetic algorithm (NSGA-II)~\cite{NSGA2} and reference-point based non-dominated sorting genetic algorithm (NSGA-III)~\cite{NSGA3}. As a result, two Pareto-based techniques are introduced here, namely, EnNSGAII and EnNSGAIII.
	\item We provided an extensive set of experiments for validating the algorithms.
\end{enumerate}

This paper is organised as follows. Section~\ref{Sec:challenges} explains briefly some challenges in the paper, while Section~\ref{Sec:background} introduces background knowledge. Section~\ref{Sec:Algorithms} explains the metaheueristic algorithms used in the paper. The proposed algorithms are introduced in Section~\ref{Sec:proposed}, whereas we provide a set of extensive experiments in Section~\ref{Sec:exp}. Finally, the paper is concluded. 



\section{Key Challenge}
\label{Sec:challenges}

One of the main challenges in calculating the energy consumption of an application is figuring out how to do it. Some research uses hardware devices for this purpose~\cite{Earmo}, which is hard to set up. They calculate the energy consumption of the battery and not an android application. Some others try to estimate the energy profile of an android application, which is not straightforward to do as well~\cite{Petra}.

This section investigates the effect of image file size and image quality on energy consumption. To this end, we used an energy profiler in the Android Studio software~\cite{android} and a plot digitiser~\cite{plot_digitiser} to estimate energy consumption. Plot digitisers are tools to convert a specific curve to digitised numbers. To this end, first, the digitiser should be calibrated for the curve (here between 0 and 2000). In other words, we should specify the minimum  and maximum values on the y-axis. In this case, the minimum and maximum values are not critical since we only need a comparison between the results (and not an exact number for energy consumption). 

\begin{figure}
	\centering
	\includegraphics[width=.95\columnwidth]{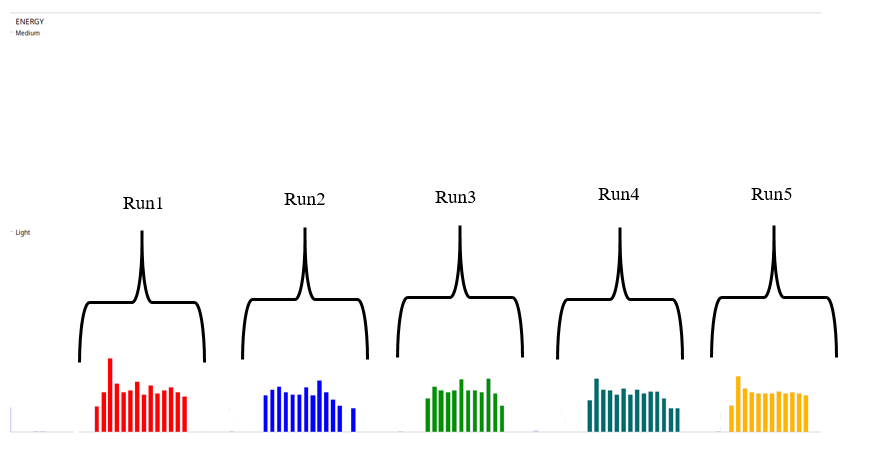}
	\caption{Energy profiler for an image loading program in 5 independent runs.} 
	\label{fig:profiler}
\end{figure}

To design the experiment, we compress an image at different levels (90\%, 70\%, 50\%, 30\% and 10\%), and the energy consumption (EC) for each image is calculated. To this end, an image loading program is written, located in a loop with 1000000 iterations and then, the energy profile is achieved for this during five independent runs (Figure~\ref{fig:profiler}). Then, the energy profiler yielded is fed to the plot digitiser to convert it to digit numbers, and for each run, the total energy consumption is estimated. Finally, the average over five runs is obtained as the EC measure. 
The size and quality (based on PSNR) are also reported for each image. The results can be seen in Table~\ref{ECI_image_size}. From the table, we can conclude that:
\begin{enumerate}
	\item By decreasing the image compression level, the EC is decreased as well.
	\item By decreasing the image compression level, as expected, the file size is decreased as well.
	\item By decreasing the image compression level, as expected, the image quality  deteriorated.
	\item By decreasing the image size, the image quality is also reduced.
\end{enumerate}

All in all, reducing the compression level reduces the file size, image quality, and energy consumption, while the developer tends to use higher image quality and smaller file size. Therefore, these two objectives, image quality and image size, are in conflict. 

The correlation between image size and image quality (based on the information available in Table~\ref{ECI_image_size}) is demonstrated in Table~\ref{ECI_corr} to show this contradiction. This table clearly verifies a conflict between higher image quality and smaller file size since the correlation is a high positive number, close to 1. 
\begin{table}[]
	\centering
	\caption{Effect of image size and image quality on the energy consumption.}
		\label{ECI_image_size}
	\begin{tabular}{l|cccc}
		\hline
			Level               & EC             & File Size(Mb)                            & PSNR                             \\ \hline
		Original               & 2769.52             & 3.07                            & inf                             \\
		90                     & 2734.73             & 1.90                            & 38.6913                   \\
		70                     & 2682.73             & 1.50                            & 37.5774                    \\
		50                     & 2638.61             & 0.98                            & 35.1707                    \\
		30                     & 2566.47             & 0.66                            & 33.0362                    \\
		10                     & 2511.69             & 0.27                            & 28.3434                    \\ \hline
	\end{tabular}
\end{table}

\begin{table}[]
	\centering
	\caption{Correlation between EC criterion and file size with other.}
	\label{ECI_corr}
	\begin{tabular}{l|ccc}
		Correlation & File size & PSNR    \\ \hline
		EC          & 0.9433     & 0.9754  \\
		File size  & 1          & 0.9615  \\ \hline
	\end{tabular}
\end{table}

In conclusion, file size and image quality can be considered proxies of energy consumption. While developers tend towards smaller file sizes and higher image quality, these two objectives conflict with each other since higher image quality will increase the file size and energy consumption. As a result, it is necessary to strike a balance between image quality and file size.   
\section{ Preliminaries}
\label{Sec:background}

\subsection{Multi-objective Optimisation}
Multi-objective optimisation (MO) is the process of finding the minimum or maximum of two conflicting objective functions. Without loss of generality, a multi-objective optimisation problem (MOP), formally, can be stated as a minimisation problem as

\begin{equation}
\label{eq:1}
\begin{aligned}
Minimise \quad F(x)= (f_{1}, f_{2}, ..., f_{M} ) \\
subject \quad to \quad x \in \Omega
\end{aligned}
\end{equation}

where $\Omega$ is the decision space, and $F:\Omega \rightarrow$ $\mathbb{R}^M$ is the objective function in which $M$ is the number of different real-valued objective functions , and $\mathbb{R}^M$ shows the objective space.

There are two general metaheuristic approaches for tackling multi-objective optimisation, scalarisation and Pareto-based approaches~\cite{Survey_MOO}. Scalarisation approaches solve a multi-objective problem by converting it into a single-objective problem, while Pareto-based techniques find a set of optimal solutions. 

Scalarisation approaches incorporate multi-objective functions into one single scalar objective function as
\begin{equation}
\label{eq:scalar}
G(x)=w_{1}f_{1}(x)+w_{2}f_{2}(x)+...+w_{M}f_{M}(x)
\end{equation}

The real-valued positive weights, $w_{i}, i=1,2,...,M$, indicate the performance priority. A larger weight shows that the corresponding objective function has a higher priority than the objective function with a smaller weight. When the priority of objective functions is not clear in advance, one of the most common methods is to use Equal Weights~\cite{Survey_MOO}, in which the weights are given by  
\begin{equation}
\label{eq:scalar}
w_{i}=\frac{1}{M}
\end{equation}
where $i=1,2,...,M$.

After scalarisation, all single-solution-based metaheuristic algorithms such as GA~\cite{GA_Main_Ref}, PSO~\cite{PSO_Main_Paper02}, and DE~\cite{DE_Original} can be used to find the optimal solution.

There is usually no single solution that can simultaneously minimise all the objective functions since the objectives are inherently competing. To tackle this, a set of optimal solutions, called \textit{Pareto optimal solutions}, can be defined, with corresponding localisation in the objective space called the \textit{Pareto front}. 

In single-objective optimisation, the superiority of one solution over another can be easily obtained by comparing the objective functions, while the quality of a solution can be achieved by the concept of dominance in multi-objective optimisation.

\textbf{Definition 1} (Pareto dominance). A solution $x_{1}$ dominates another solution $x_{2}$ (denoted by $x_{1} \prec x_{2}$) if and only if:

\begin{enumerate}
    \item $\forall i \in \{1,2,...,M\} : f_{i} (x_{1}) \leqslant f_{i} (x_{2})$, where $M$ is the number of objective functions. In other words, in all objective functions, solution $x_{1}$ should not be worse than $x_{2}$.
    \item $\exists j \in \{1,2,...,M\} : f_{i} (x_{2}) < f_{i} (x_{1}) $; meaning that solution $x_{1}$ is strictly superior to solution $x_{2}$ in at least one objective function.
\end{enumerate}

\textbf{Definition 2} (Non-dominated solution). A solution $x_{1}$ is called Pareto optimal solution or non-dominated solution if it is not dominated by other solutions in the whole search space. It can be mathematically defined as 
\begin{equation}
\label{eq:pareto_optimality}
\nexists x_{2} \in X : x_{2} \prec x_{1} 
\end{equation}

\begin{figure}
	\centering
	\includegraphics[width=.9\columnwidth]{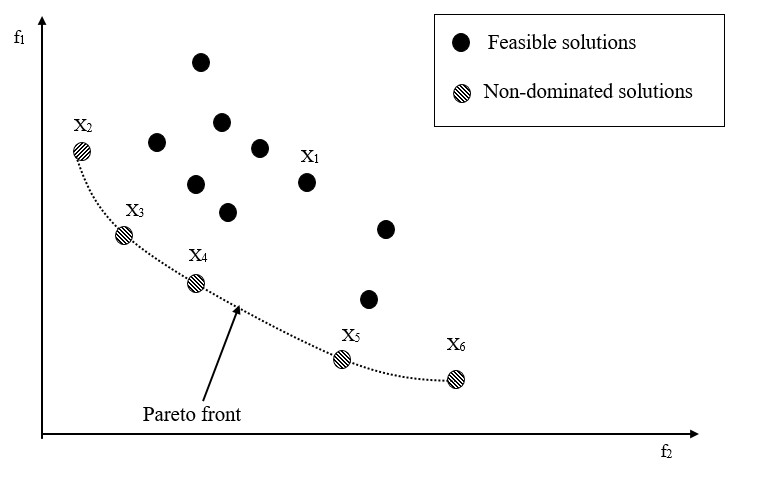}
	\caption{Non-dominated solutions in a bi-objective optimisation problem.} 
	\label{fig:Pareto}
\end{figure}

Figure~\ref{fig:Pareto} indicates non-dominated solutions among other solutions in a bi-objective minimisation problem. $f_{1}$ and $f_{2}$ are two conflicting objectives which should be minimised simultaneously. From the figure, $x_{3}$ has a lower value than $x_{1}$ in both objective functions. Therefore, we can say that $x_{1}$ is dominated by $x_{3}$. In other words, $x_{3}$ is a non-dominated solution. In addition, $x_{2}$, $x_{4}$, $x_{5}$ and $x_{6}$ are also non-dominated solutions since there is no other solution that dominates them in both objective functions.

\textbf{Definition 3} (Pareto front). The set of all non-dominated solutions is called  Pareto optimal set (PS), which is stated as
\begin{equation}
\label{eq:pareto_optimality}
PS= \{u \in X|\nexists v \in X, u \prec v\}
\end{equation}

The Pareto front (PF) corresponds to the Pareto optimal set in the objective space, and is denoted as
\begin{equation}
\label{eq:pareto_front}
PF= \{F(x)|x \in PS\}
\end{equation}

\subsection{The JPEG Image Compression}
\label{JPEG_Image}
\begin{figure}
	\centering
	\includegraphics[width=.95\columnwidth]{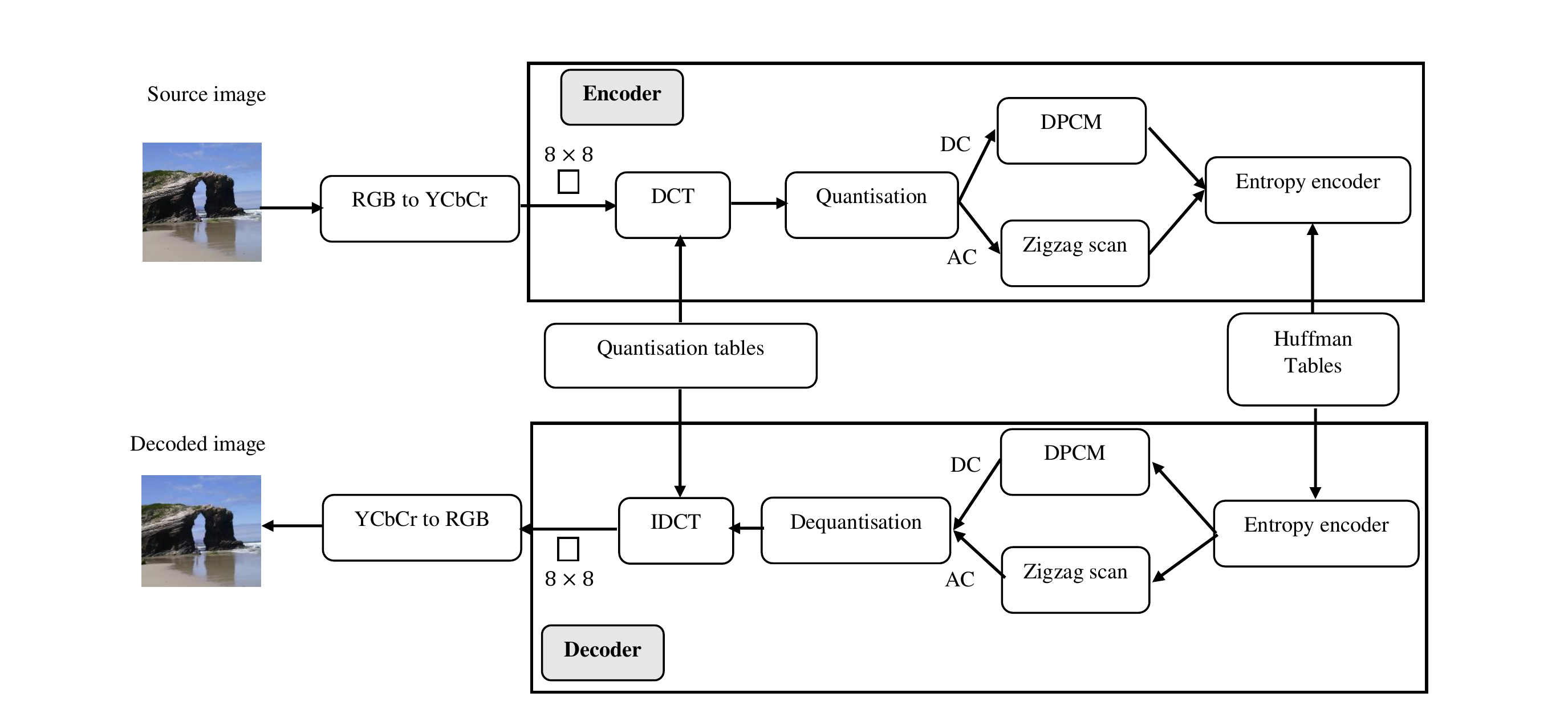}
	\caption{The main components of the JPEG image compression.} 
	\label{fig:Jpeg}
\end{figure}

Figure~\ref{fig:Jpeg} shows the main components of JPEG image compression. The encoder is responsible for converting the original image into the JPEG compression variant of the original image, while the reverse task is carried out by the decoder. In the following, we explain the main components in more detail. 
\subsubsection{DCT and IDCT Components}
The source image is first divided into $ 8 \times 8$ blocks. Then, the values of the blocks are shifted from $[0,2^{p}-1]$ to $[-2^{p-1},2^{p-1}-1]$, in which $p$ is the number of bits per pixel (in the baseline JPEG compression, $p=8$). Each block of $8 \times 8$ pixels can be seen as a vector with a size of $64 \times 1$, which should be fed into the Discrete Cosine Transform (DCT)~\cite{DCT_Journal} component. The DCT block decomposes the input signal into 64 basis-signal amplitudes, called DCT coefficients. Mathematically, the DCT can be expressed as
\begin{equation}
\label{eq:DCT}
F(u,v)=\frac{1}{4}c_{u}c_{v}  \left[ \sum_{x=0}^{7} \sum_{y=0}^{7} f(x,y) \cos \left( \frac{(2x+1)u\pi}{16} \right) \cos \left(\frac{(2y+1)v\pi}{16} \right) \right]
\end{equation}
where
\begin{equation}
c_{r}=\begin{cases}
\frac{1}{\sqrt{2}} & r=0 \\
1 & r>0
\end{cases} ,
\end{equation}
The coefficient corresponding to $u,v=0$ is known as the DC coefficient, while the remaining 63 coefficients are the AC coefficients. 

Inverse DCT (IDCT) is the reverse of DCT component to reconstruct the original image, which is stated as
\begin{equation}
\label{eq:DCT}
F(x,y)=\frac{1}{4}c_{u}c_{v} \left[ \sum_{u=0}^{7} \sum_{v=0}^{7} f(u,v)  \cos  \left(\frac{(2x+1)u\pi}{16} \right) \cos \left(\frac{(2y+1)v\pi}{16} \right) \right]
\end{equation}

In the absence of the quantisation step, the original 64-point signal is precisely restored.

\subsubsection{The Quantisation and Dequantisation Components}

The quantisation step works based on a 64-element quantisation table, which should be known in advance. Each table entry defines the step size of the quantiser for its related DCT coefficient and belongs to [1,255]. Quantisation aims to achieve compression while maintaining image quality by removing information that is not visually important.

The quantisation component is defined as
\begin{equation}
L(u,v)= round\left(  \frac{F(u,v)}{Q(u,v)}\right),
\end{equation}
where $L(u,v)$ are the quantised DCT coefficients, $F(u,v)$ are the DCT coefficients, $Q(u,v)$ indicates the corresponding element of the quantisation table, and $round(x)$ is the closets integer number to $x$. It is worth mentioning that the larger the value of $Q(u,v)$, the larger the information loss.

The de-quantisation component of the decoder reverses the quantisation process to recreate a rough estimate of $F(u,v)$ from $L(u,v)$ as 
\begin{equation}
\bar{F}(u,v)=L(u,v) \times Q(u,v),
\end{equation}

This step plays a crucial role in the process of JPEG compression since the quantisation table generates a loss of information. Thus, it is necessary to establish the quantisation table to strike a compromise between compression effectiveness and reconstructed image quality.
\subsubsection{Symbol Coding }
The 63 AC coefficients of the $8 \times 8$ block are handled independently from the DC coefficient after quantisation. The Differential Pulse Code Modulation (DPCM) is used to encode the DC coefficient as 
\begin{equation}
DIFF_{i}=DC_{i}-DC_{i-1}
\end{equation}
where $DC_{i}$ and $DC_{i-1}$ are the DC coefficients for the current $8 \times 8$ block and the prior $8 \times 8$ block, respectively.

In order to format the quantised 63 AC coefficients for entropy coding, a zigzag scan~\cite{JPEG_Zigzag} can be used. After the zigzag scan, the AC coefficients show diminishing variances and rising spatial frequencies.
\subsubsection{Entropy Coding }
After the quantisation process, there are often a few nonzero and several zero-valued DCT coefficients. Entropy coding's goal is to compress the quantised DCT coefficients by making use of their statistical properties. The baseline technique used by JPEG is the Huffman coding, which employs two DC and two AC Huffman tables for the luminance and chrominance DCT coefficients, respectively~\cite{JPEG_Zigzag}. 

\section{Algorithms}
\label{Sec:Algorithms}

Due to the introduction of a vast and varied range of metaheuristic techniques in the literature, it is evident that we cannot analyse all of them. Also, the main focus of this paper is not benchmarking all algorithms but introducing a general strategy for multi-objective JPEG image compression. Therefore, for our study, we have chosen a variety of state-of-the-art algorithms. In the following, we briefly outline the selected algorithms, while the cited publications are referred to for further details.
\subsection{Scalarisation Methods}
\label{sec:scalar}

\begin{compactitem}
	\item
 Genetic algorithm (GA)~\cite{GA_Main_Ref} is the oldest metaheuristic algorithm, and includes three significant operators: selection, crossover, and mutation. The selection operator is responsible for selecting candidate solutions who contribute to the next generation's population. The information from the parents is integrated into the crossover operator, while random modifications are made to one or more components of a potential solution in the mutation operator. Based on the "survival of the fittest" premise, solutions are transferred from one iteration to the next.
	
	\item
	Differential evolution (DE)~\cite{DE_Original} is another metaheuristic algorithm including three main operators, mutation, crossover, and selection. Mutation creates candidate solutions based on the differences among candidate solutions as
	\begin{equation}
	\label{Eq:DE}
	v_{i}=x_{r1} + SF (x_{r2}-x_{r3}) ,
	\end{equation} 
	where $SF$ signifies a scaling factor, and $x_{r1}$, $x_{r2}$, and $x_{r3}$ are three distinct randomly selected candidate solutions from the current population, and $v_{i}$ is called a mutant vector. Crossover is responsible for integrating the mutant vector with a target vector selected from the current population. Eventually, a candidate solution is selected by a selection operator depending on its quality.
	
	\item
	Particle swarm optimisation~\cite{PSO_Main_Paper} is a swarm-based optimisation approach, and its updating scheme is based on the best position found for each candidate solution and a global best position. The velocity vector of a particle is updated as   
	\begin{equation}
	\label{Eq:vel}
	v_{t+1}= \omega v_{t}+c_{1} r_{1} (p_{t}-x_{t})+c_{2} r_{2} (g_{t}-x_{t}) ,
	\end{equation} 
	where $t$ shows the current iteration, $x_{t}$ is the current position, $r_{1}$ and $r_{2}$ are random numbers generated from a uniform distribution in the range of $[0,1]$, $p_{t}$ is the personal best position, and $g_{t}$ indicates the global best position. Then, a candidate solution is updated as
		\begin{equation}
	\label{Eq:pop}
	x_{t+1}=x_{t} +v_{t+1} ,
	\end{equation}
	\item Evolutionary strategy~\cite{ES_main_paper} is a metaheuristic algorithm where each offspring is generated based on a Gaussian random number as
		\begin{equation}
	\label{Eq:ES}
	x_{new}= x_{old}+N(0,\sigma^{2}) ,
	\end{equation} 
	where $N(0,\sigma^{2})$ is a  Gaussian random number with mean 0 and variance $\sigma^{2}$. Then, competition should be done for each individual and finally, the best individuals transfer to the next generation.
	\item Pattern search~\cite{PS_main_paper} is a simple yet effective optimisation algorithm that, in an iterative manner, combines exploratory and pattern moves to find the best solution to a problem. The exploratory move tries one direction, and if that doesn't work, it tries the other. In particular, it generates a new solution as 
	\begin{equation}
	\label{Eq:PS1}
	x^{+}= x+\rho ,
	\end{equation} 
	where $x^{+}$ is the new solution based on the current solution $x$, and  $\rho$ is called the step size or exploratory radius. If this move can not improve the current solution, it attempts another direction
	\begin{equation}
	\label{Eq:PS2}
	x^{+}= x-\rho ,
	\end{equation}
	If the moves in all directions fail, then the radius is halved.  
\end{compactitem}	

\subsection{Pareto-based Techniques}
	
\subsubsection{Non-dominated Sorting Genetic Algorithm II}
\label{sec:nsga2}
The Non-Dominated Sorting Genetic Algorithm (NSGA-II)~\cite{NSGA2} is one of the state-of-the-art approaches for Pareto-based multi-objective optimisation. NSGA-II is based on four basic operators, including, Non-Dominated Sorting, Elite Preserving Operator, Crowding Distance, and Selection Operator, which are described below in more detail.

\begin{figure}
	\centering
	\includegraphics[width=.8\columnwidth]{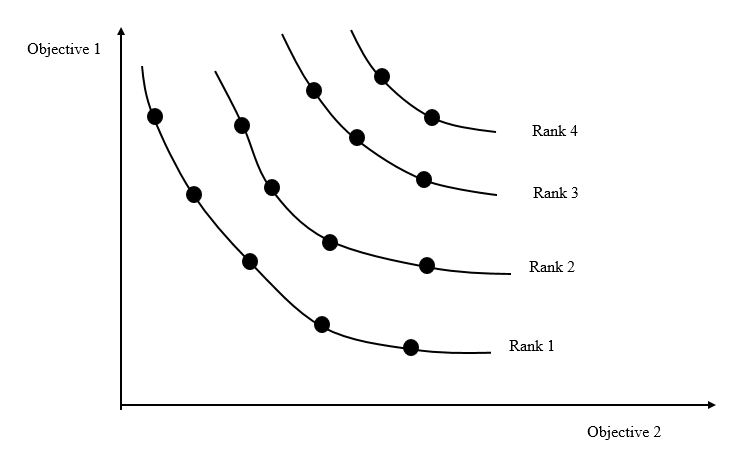}
	\caption{Non-dominated sorting procedure.} 
	\label{fig:NDS}
\end{figure}

\textbf{Non-Dominated Sorting:}
The notion of Pareto dominance is used in this process to sort the population members. In the first step,  the non-dominated members of the initial population are assigned to the first rank. These top-ranked individuals are subsequently put in the first front and eliminated from the current population. The remaining population members are then subjected to the non-dominated sorting technique. The remaining population's non-dominated individuals are given the second rank and positioned in the second front. This procedure continues until all population members are distributed across various fronts in accordance with their rankings, as seen in Figure~\ref{fig:NDS}.

\textbf{Elite Preserving Operator:}
Elite solutions are maintained by being immediately passed on to the next generation as part of an elite preservation strategy. In other words, the non-dominated solutions identified in each generation transfer to the next generations until some solutions dominate them.

\begin{figure}
	\centering
	\includegraphics[width=.8\columnwidth]{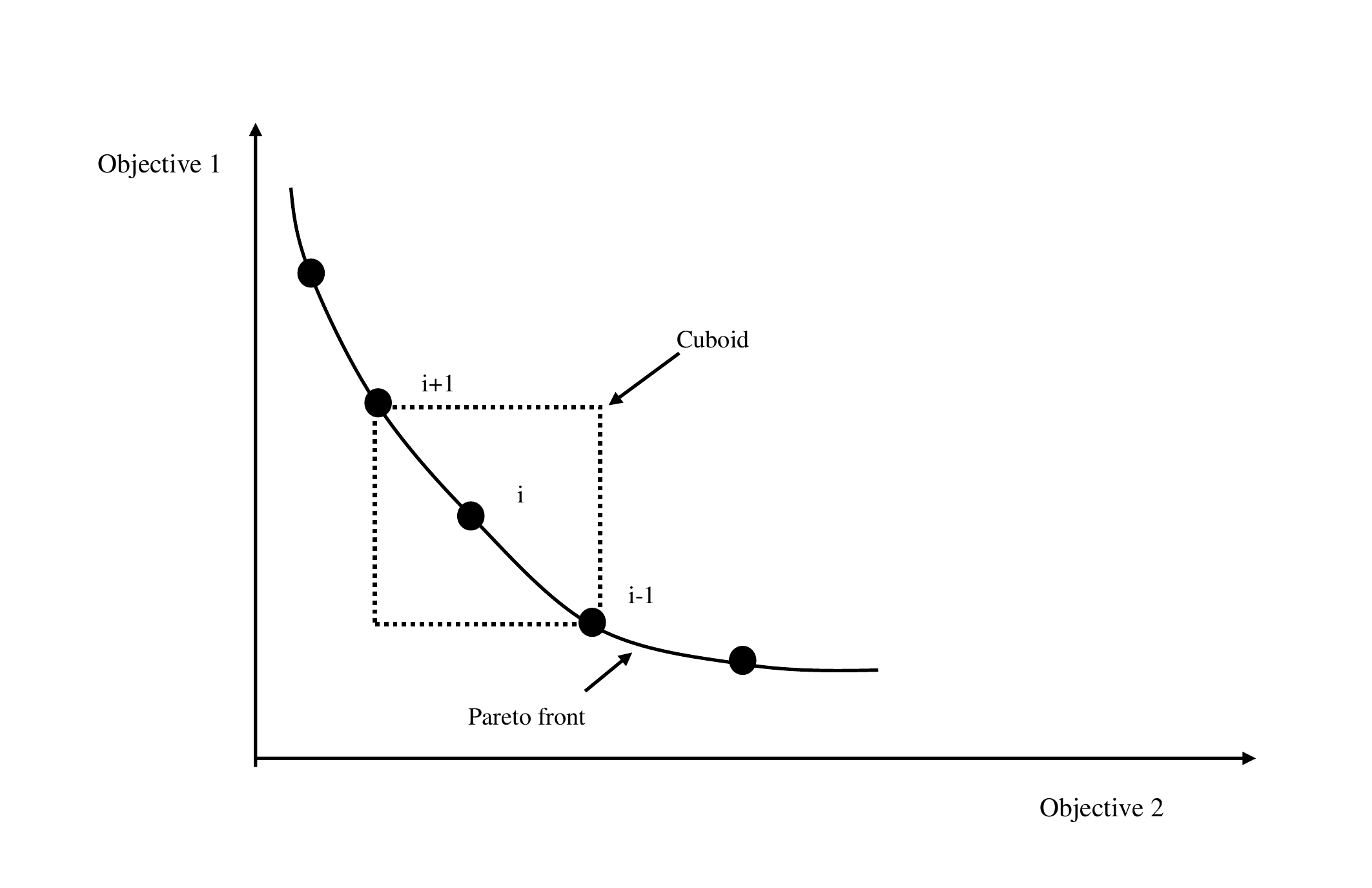}
	\caption{Crowding distance.} 
	\label{fig:crowding}
\end{figure}

\textbf{Crowding Distance:}
The crowding distance determines the density of solutions around a specific solution. It is the average distance between two solutions along each of the objectives on each side of the solution. When two solutions with varying crowding distances are compared, the solution with the greater crowding distance is assumed to be present in a less crowded area. The crowded distance of the $i$-th solution is computed based on the average side-length of the cuboid (Figure~\ref{fig:crowding}). Mathematically, the crowding distance is defined as
		\begin{equation}
\label{Eq:CD}
CD_{i}= \sum_{j=1}^{k} \frac{f_{j}^{i+1}-f_{j}^{i-1}}{f_{j}^{max}-f_{j}^{min}} ,
\end{equation} 
where $f_{j}^{i}$ shows the $j$-th value of an objective function for the $i$-th solution, $f_{j}^{max}$ and $f_{j}^{min}$ signify the maximum and minimum values of $j$-th objective function among the current population, and $k$ is the number of objective functions. 

\textbf{Selection Operator:}
A crowded tournament selection operator is used to choose the population for the next generation. This operator selects the population based on the rank of the population members and the crowding distances between them. The following rules apply when choosing one of two population members to represent the next generation: 1) If the two population members are of different ranks, the higher rank one is chosen; and 2) If the two population members are of the same rank, the member with the greater crowding distance is chosen.

\textbf{Procedure:}
The NSGA-II algorithm starts by creating an initial population $P_{t}$ of size $N$. Following crossover and mutation operations on the population, $P_{t}$, a new population, $Q_{t}$, is produced. Then, the non-dominated sorting operation is carried out on the new population, $R_{t}$, created by combining the populations $P_{t}$ and $Q_{t}$. The $R_{t}$ population members are then divided into several fronts based on their non-dominance levels.

\begin{figure}
	\centering
	\includegraphics[width=1\columnwidth]{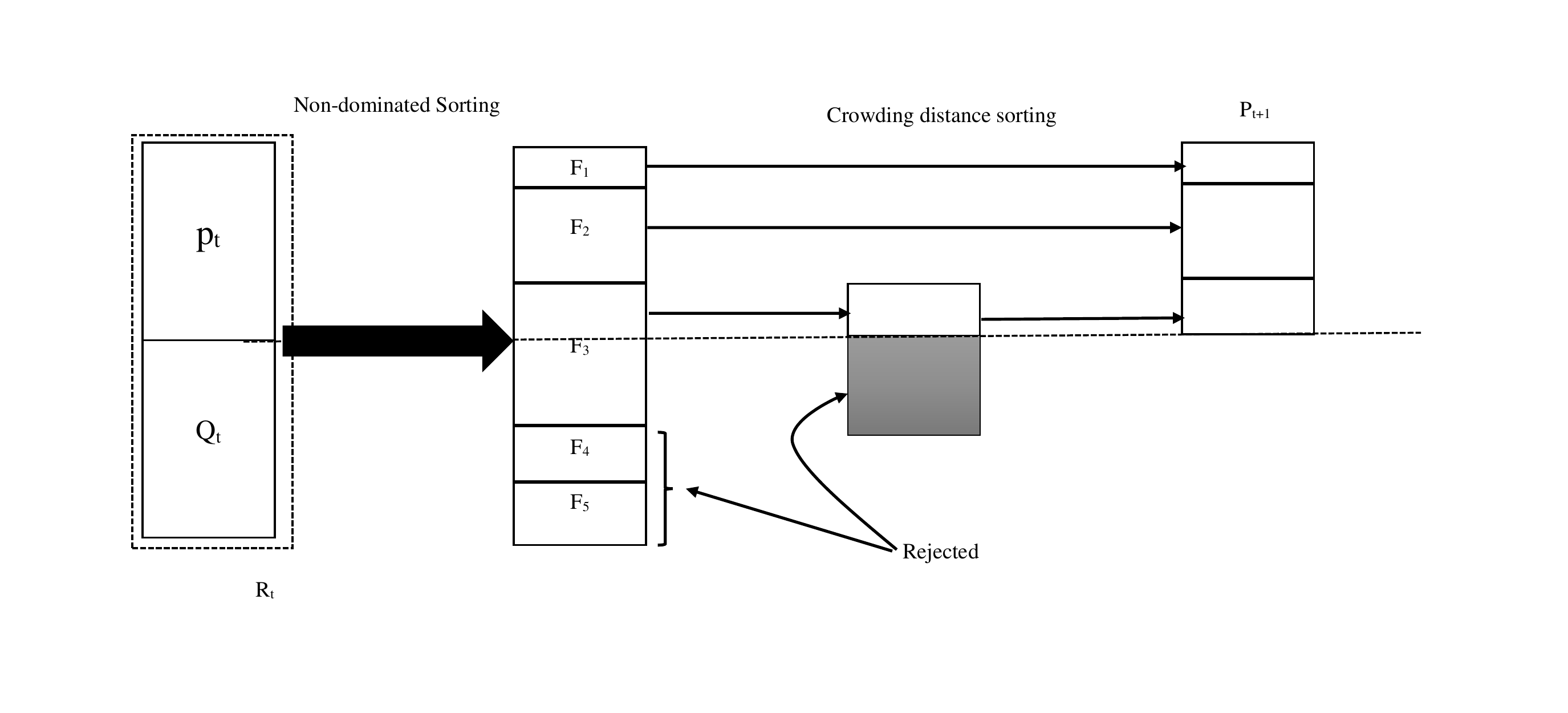}
	\caption{The NSGAII procedure.} 
	\label{fig:nsga2}
\end{figure}

The next step is to choose $N$ candidate solutions from $R_{t}$ in order to produce $P_{t+1}$. If the size of the first front is greater than or equal to $N$, only $N$ members are chosen from the least crowded area of the first front to create $P_{t+1}$. The members of the first front are directly moved to the next generation if the size of the first front is more than $N$, and the remaining members are taken from the second front's least crowded area and added to $P_{t+1}$. The process is repeated for the subsequent fronts until the size of $P_{t+1}$ equals $N$, if the size of $P_{t+1}$ is still less than $N$. Following the same process, the populations $P_{t+2}$, $P_{t+3}$, $P_{t+4}$,... for subsequent generations are created until the stopping criterion is not met. Figure~\ref{fig:nsga2} shows the NSGAII procedure visually.

\subsection{Reference-point Based Non-dominated Sorting Genetic Algorithm }

The basic framework of Reference-point Based Non-dominated Sorting Genetic Algorithm (NSGA-III)~\cite{NSGA3} is similar to NSGA-II, but with significant modifications to its selection process. Unlike NSGA-II, NSGA-III adaptively updates several widely used reference points, which aids in maintaining diversity among population members. 

As previously mentioned, the NSGA-III employs a pre-defined set of reference points to guarantee diversity in the solutions produced. The standard NSGA-III algorithm benefits from Das and Dennis’s~\cite{ref_points} approach which assigns points to a normalised hyper-plane. The total number of reference points ($H$) in an $M$ objective problem by $P$ division can be calculated as 
		\begin{equation}
\label{Eq:Das}
H = \binom{M+P-1}{P}
\end{equation}

\begin{figure}
	\centering
	\includegraphics[width=0.8\columnwidth]{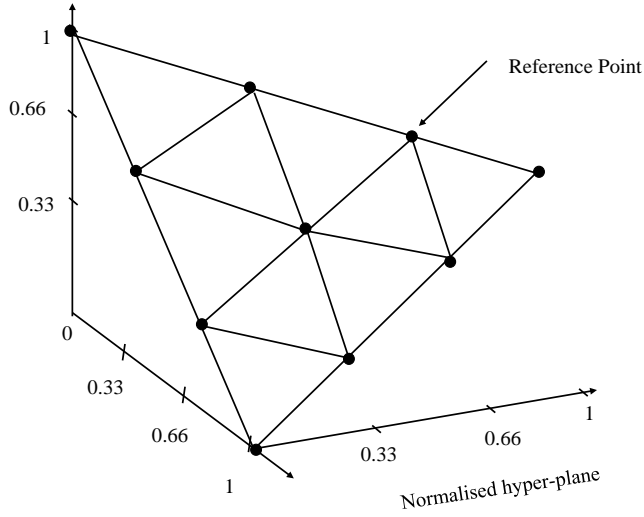}
	\caption{ 3D plot of the 10 reference points with p = 3.} 
	\label{fig:ref}
\end{figure}

For instance, in a problem with three objectives ($M=3$), the reference points are made on a triangle whose apex is at (1, 0, 0), (0, 1, 0), and (0,0,1). For each objective axis, four divisions ($P=3$) will result, and therefore a total of 10 reference points (Figure~\ref{fig:ref}).

NSGA-II utilises the crowding distance to pick the remaining members after non-dominated sorting, while with NSGA-III, the reference points are used to select the remaining members. 
To do this, the range of objective values and reference points are first normalised to be the same. Afterwards, the orthogonal distance between each reference line and a member of $P_{t}$ is calculated. The reference point with the shortest orthogonal distance is then used to associate the member with.

The number of individuals who are connected to each reference point, known as the niche count for each reference point, is then computed for further analysis. Then, the reference point with the lowest niche point is found and the member from
the last front that is associated with it should be included in the final population. It is important to note that a reference point need not have any population members linked with it and may have one or more related population members. For the $j$-th reference point, the number of population members that are associated with each reference point is counted and denoted as niche count ($\rho_{j}$). NSGA-III employs a niche-preserving operator as follows. First, a reference point set, $J_{min}={j:argmin_{j} \rho_{j}}$ is defined, including a minimum $\rho_{j}$. When there are several of these reference points, one ($j^{*} \in j_{min}$) is selected at random. If $\rho_{j^{*}}=0$ (meaning that there is no associated member to the reference point $j$), two scenarios can happen. First, the reference point $j$ is already connected to one or more of the members in front of $F_{1}$. In this case, the one closest to the reference line perpendicularly is added to $P_{t+1}$. Second, the front $F_{1}$ does not have any members linked to the reference point. In this case, the reference point is not taken into account anymore for the current generation.

In the case of $\rho_{j} \geq 1$ (indicating that one member associated with the reference point exists in $P_{t} /F_{L}$ ), a randomly selected number from front $F_{L}$ that is associated with the reference point $ \rho_{j}$ is added to $P_{t+1}$. After updating the niche counts, the process is repeated $K$ times in total to bring $P_{t+1}$'s population size to $N$.

\section{Proposed Methods}
\label{Sec:proposed}
This paper proposes a general strategy for multi-objective optimisation of JPEG implementation. As a result, it can be used with any optimisation algorithm. To this end, we embed our proposed scheme into five scalarisation algorithms, GA, DE, PSO, ES, and PS, and two Pareto-front-based algorithms, NSGA-II and NSGA-III. First, we define the solution representation and objective function in the following. Then, by embedding the proposed strategy within seven backbone optimisation algorithms, we obtain seven new algorithms, namely EnMOGA, EnMODE, EnMOPSO, EnMOES, EnMOPS, EnNSGAII, and EnNSGAIII, respectively. 

\subsection{Solution Representation}
\label{sec:rep}
Our proposed algorithm aims to find multi-objective optimal quantisation tables, including luminance quantisation table (LQT) and chrominance quantisation table (CQT). To this end, each 8-by-8 quantisation table is reshaped to a 1-by-64 vector, and then both are concatenated. Therefore, the representation proposed in this paper is a vector of dimension 128 as
 	\begin{equation}
 x=[LQT_{1,1},...,LQT_{8,8},...,CQT_{1,1},...CQT_{8,8}]
 \end{equation}   
 
 where $LQT_{i,j}$ and $CQT_{i,j}$ show the corresponding element in the location $(i,j)$ in the LQT matrix and the CQT matrices. In other words, the first 64 entries are positive integer numbers in $[0,2^{p}-1]$ (where $p$ is the number of bits indicating a pixel, in our case $p=8$) for the LQT table, while the remaining elements are reserved for the CQT table.

\subsection{Objective Functions}
\label{sec:obj}
This paper introduces two main objective functions, file size and image quality. One is to be minimised (file size) and the other to be maximised (image quality). To this end, first, the JPEG image should be achieved using the corresponding candidate solution for a typical image. The first objective function is file size defined as
\begin{equation}
FS_{obj}=\frac{FS_{JPEG}}{FS_{org}}
\end{equation} 
where $FS_{JPEG}$ is the file size for image after JPEG compression process, and $FS_{org}$ is the image size for the original image. Lower $FS_{obj}$ shows a higher capability in the compression process.

The second objective function is Peak Signal to Noise (PSNR), as one of the most common measures for assessing image quality, which is computed as  
\begin{equation}
PSNR = 20 \log_{10}(255/RMSE) ,
\end{equation} 
where $RMSE$ is the root mean squared error which is calculated as 
\begin{equation}
RMSE = \sqrt{\dfrac{\sum_{i=1}^{M}\sum_{j=1}^{N}(I(i,j)-\hat{I}(i,j))^{2}}{MN}} ,
\end{equation} 
where $M$ and $N$ are the image dimensions, and $I$ and $\hat{I}$ are the original and the compressed images. A higher PNSR value indicates better performance.

Scalarisation approaches integrate the multi-objective functions into one objective function. Therefore, the objective function for the scalarisation methods is expressed as
 \begin{equation}
 F(x) =w_{1}.FS_{obj}+\frac{w_{2}}{PSNR}
 \end{equation} 
 where $w_{1}$ and $w_{2}$ are two used-defined parameters, indicating the importance of each objective function.
 
Pareto-based approaches can work on our two objective functions independently, so there is no need to combine two objective functions. As a result, the two objective functions for Pareto-based approaches are $FS_{obj}$ and $\frac{1}{PSNR}$.  

\subsection{Embedding Within Scalarisation Approaches}

\subsubsection{EnMOGA Algorithm}
\label{sec:EnMOGA}
The EnMOGA begins with forming a random initial population from a uniform distribution. Over various generations, new populations are created by applying crossover, mutation, and selection operators. The pseudo-code of the EnMOGA algorithm is given in Algorithm~\ref{alg:EnGA-MO}, while the components are briefly explained below. 

\textbf{Selection:} We use tournament selection, which promotes quicker convergence. In tournament selection, the top candidate solutions are chosen from a random subset of the population for each tournament. The size of the tournament is determined by the number of participants in each tournament.

\textbf{Crossover:} We use Simulated Binary Crossover (SBX)~\cite{SBX_PM} for the crossover operator.  A binary notation can express real values, and then a point crossover can be performed. By using a probability distribution model of the binary crossover, SBX replicated this process. SBX benefits from two leading parameters, including the probability of a crossover and the distribution index ($\eta$).

\textbf{Mutation:} Polynomial Mutation~\cite{SBX_PM} is used in this paper, which follows the same probability distribution as the SBX operator in the parent’s vicinity. It also has the same parameters as the SBX operator.

\begin{algorithm}[!htbp]
	\caption{EnGAMO algorithm in the form of pseudo-code.}
	\label{alg:EnGA-MO}
	\begin{algorithmic}[1]
		\State // $L$/$U$: lower/upper bound;  $N_{pop}$: population size; $N_{var}$: number of variables; $NFE_{\max}$:  maximum number of function evaluations; $prob$: the probability of the crossover, $\eta$: the distribution index
		\State Initialise population of $N_{pop}$ candidate solutions using the representation introduced in Section~\ref{sec:rep}.
		\State Calculate objective function values (OFV) of all candidate solutions (Section~\ref{sec:obj}).
		\State $x^{*}$ = best candidate solution in the initial population
		\State $NFE=N_{pop}$
		\State $iter=0$
		\While {$NFE<=NFE_{\max}$}
		\State $iter=iter+1$
		\State Perform Tournament selection (Section~\ref{sec:EnMOGA}).	
		\State Perform SBX crossover (Section~\ref{sec:EnMOGA}).
		\State Perform Polynomial mutation (Section~\ref{sec:EnMOGA}).
		\State Calculate objective function values of all new candidate solutions (Section~\ref{sec:obj}).
		\State Replace the old population by the new one.
		\State  $x^{+}$ = best candidate solution in the current population
		\If {OFV of $x^{+}$ $<$ OFV of $x^{*}$}
		\State $x^{*} =x^{+}$
		\EndIf
		\State $NFE = NFE + N_{pop}$	
		\EndWhile
	\end{algorithmic}
\end{algorithm}

\subsubsection{EnMODE Algorithm}
\label{Sec:EnMODE}
Since EnMODE is a population-based metaheuristic, it is started with a random initial population. It has three main operators, including, mutation, crossover, and selection. For them, we used the standard operators, described in Section~\ref{sec:scalar}. Also, EnMODE benefits from Dither~\cite{DE_Dither}, a deterministic scheme of randomisation of the scale factor $SF$ (introduced in Section~\ref{sec:scalar}). Dither proposes selecting $SF$ from the interval [0.5, 1.0] randomly for each generation. The pseudo-code of EnMODE is given in Algorithm~\ref{alg:EnDE-MO}. 

\begin{algorithm}[t!]
	\caption{EnMODE algorithm in the form of pseudo-code.}
	\label{alg:EnDE-MO}
	\begin{algorithmic}[1]
		\State // $L$/$U$: lower/upper bound;  $N_{pop}$: population size; $N_{var}$: number of variables; $NFE_{\max}$:  maximum number of function evaluations; $CR$: crossover rate.
		\State Initialise population of $N_{pop}$ candidate solutions using the representation introduced in Section~\ref{sec:rep}.
		\State Calculate objective function values of all candidate solutions (Section~\ref{sec:obj}).
		\State $x^{*}$ = best candidate solution in the initial population
		\State $NFE=N_{pop}$
		\State $iter=0$
		\While {$NFE<=NFE_{\max}$}
		\State $iter=iter+1$
		\State Perform Dither operation (Section~\ref{Sec:EnMODE}).
		\State Perform Mutation operator (Section~\ref{sec:scalar}).	
		\State Perform Crossover operator (Section~\ref{sec:scalar}).
		\State Calculate objective function values of all new candidate solutions (Section~\ref{sec:obj}).
		\State Perform Selection operator(Section~\ref{sec:scalar}).
		\State  $x^{*}$ = best candidate solution in the current population	
		\EndWhile
	\end{algorithmic}
\end{algorithm}

\subsubsection{EnMOPSO Algorithm}
The EnMOPSO is based on the PSO algorithm. The updating strategy used here is similar to the standard PSO algorithm introduced in Section~\ref{sec:scalar}. Standard PSO uses two parameters, $c_{1}$ and $c_{2}$. Here, both parameters are updated based on the way proposed in \cite{Adaptive_PSO}. To this end, PSO has been placed in 4 states, including convergence, exploitation, exploration, and jumping out. In each state, one of the following operations should be performed.
\begin{enumerate}
	\item Increasing $c_{1}$ and decreasing $c_{2}$ in an exploration state,
	\item Increasing $c_{1}$ slightly and decreasing $c_{2}$ slightly
	in an exploitation State,
	\item Increasing $c_{1}$ slightly and increasing $c_{2}$ slightly
	in a convergence state,
	\item Decreasing $c_{1}$ and increasing $c_{2}$ in a jumping out state.
\end{enumerate}
The evolutionary states estimation process is as follows.
\begin{enumerate}
	\item Calculate the mean distance of each particle ($d_{i}$) with all other particles as
		\begin{equation}
	\label{eq:mean}
	d_{i}=\frac{1}{N-1} \sum_{j=1, j\neq i}^{N} \sqrt{ \sum_{k=1}^{D} (x_{i}^{k}-x_{j}^{k})}
	\end{equation} 
	where N and D are population size and the number of dimensions, respectively.
	
	\item Calculate the evolutionary factor as
	\begin{equation}
	\label{eq:ef}
	ef = \frac{d_{g}-d_{min}}{d_{max}-d_{min}}
	\end{equation} 
where $d_{g}$ means the distance for the global best position, and $d_{max}$ and $d_{min}$ are the maximum and minimum distances, respectively. 
	\item Classify $ef$ into one of four sets (based on the rules introduced in~\cite{Adaptive_PSO}), which represents the states of exploration, exploitation, convergence, and jumping out.
\end{enumerate}
In addition to updating $c1$ and $c2$, $\omega$ also updates based on a Sigmoid function as
	\begin{equation}
	\label{eq:omega}
\omega (ef) = \frac{1}{1+1.5e^{-2.6ef}}
\end{equation} 

The EnMOPSO algorithm in the form of Pseudo-code is given in Algorithm~\ref{alg:EnPSO-MO}. 

\begin{algorithm}[t!]
	\caption{EnMOPSO algorithm in the form of Pseudo-code.}
	\label{alg:EnPSO-MO}
	\begin{algorithmic}[1]
		\State // $L$/$U$: lower/upper bound;  $N_{pop}$: population size; $N_{var}$: number of variables; $NFE_{\max}$:  maximum number of function evaluations; $G$: maximum number of iterations.
		\State \
		\State $g=1$
		\State $NFE=N_{pop}$
		\State $iter=0$
		\State Initialise population of $N_{pop}$ candidate solutions using the representation introduced in Section~\ref{sec:rep}.
		\State Calculate objective function values of all candidate solutions (Section~\ref{sec:obj}).
		\State Initialise \textit{Gbest} as a candidate solution with the minimum value of the population.
		\State Initialise \textit{Pbest} to its initial poisiton for each candidate solution.
		\While {$NFE<=NFE_{\max}$}
		\State $iter=iter+1$
		\State Estimate the evolutionary states of the algorithm and calculate evolutionary factor using Eq.~\ref{eq:ef}
		\State Select one of 4 states, including convergence, exploitation, exploration, and jumping out to update the parameters
		\State Update $\omega$ using Eq.\ref{eq:omega}.
		\State  Calculate the particle’s velocity according to Eq.~\ref{Eq:vel}	
		\State Update particle’s position according to Eq.~\ref{Eq:pop}
		\State Calculate objective function values of all new candidate solutions (Section~\ref{sec:obj}).
		\State Update \textit{Gbest} and \textit{Pbest}.
		\State  $x^{*}$ = best candidate solution in the current population	
		\State Update $NFE$
		\EndWhile
	\end{algorithmic}
\end{algorithm}
\subsubsection{EnMOES Algorithm}
EnMOES algorithm is inspired by evolutionary strategy~\cite{ES_main_paper}, and includes two leading operators, namely mutation and selection. Mutation operator is performed using Eq.~\ref{Eq:ES}, while selection is based on objective function ranking. The EnMOES algorithm in the form of pseudo-code is given in Algorithm~\ref{alg:EnES-MO}.

\begin{algorithm}[t!]
	\caption{EnMOES algorithm in form of Pseudo-code.}
	\label{alg:EnES-MO}
	\begin{algorithmic}[1]
		\State // $L$/$U$: lower/upper bound;  $N_{pop}$: number of bids; $N_{var}$: number of variables;$NFE_{\max}$:  maximum number of function evaluations; $\sigma$: variance value for Gaussian distribution.
		\State \
		\State Initialise population of $N_{pop}$ candidate solutions using the representation introduced in Section~\ref{sec:rep}.
		\State Calculate objective function values of all candidate solutions (Section~\ref{sec:obj}).
		\State $x^{*}$ = best candidate solution in the initial population
		\State $NFE=N_{pop}$
		\State $iter=0$
		\While {$NFE<=NFE_{\max}$}
		\State $iter=iter+1$
		\State Select parents in a random manner.	
		\State Generate offspring using Eq.~\ref{Eq:ES}.
		\State Calculate objective function values of all offspring (Section~\ref{sec:obj}).
		\State Select the best candidate solutions among the combination of offspring and parents.
		\State  $x^{*}$ = best candidate solution in the current population	
		\EndWhile
	\end{algorithmic}
\end{algorithm}
 
\subsubsection{EnMOPS Algorithm}
EnMOPS works based on pattern search; therefore, it tries to find the optimal point by comparing, at each iteration, its value with a finite set of trial points. The Pseudo-code of EnPS-MO algorithm is given in Algorithm~\ref{alg:EnPS-MO}. 

\begin{algorithm}[t!]
	\caption{EnMOPS algorithm in the form of Pseudo-code.}
	\label{alg:EnPS-MO}
	\begin{algorithmic}[1]
		\State // $L$/$U$: lower/upper bound;  $N_{var}$: number of variables;$NFE_{\max}$:  maximum number of function evaluations; $\rho$: step size
		\State Generate a randomly candidate solution ($x$) using the representation introduced in Section~\ref{sec:rep}.
		\State Calculate objective function values (OFV) of the candidate solution (Section~\ref{sec:obj}).
		\State $NFE=1$
		\State $iter=0$
		\While {$NFE<=NFE_{\max}$}
		\State $iter=iter+1$
		\State Generate one trial solution ($x^{+}$) using Eq.~\ref{Eq:PS1}	
		\State Calculate objective function values of the new trial solution (Section~\ref{sec:obj}).
		\If{$x^{+}$ is better than the current solution}
		\State replace the current solution by $x^{+}$ 
		\Else
		\State Generate one trial solution ($x^{+}$) using Eq.~\ref{Eq:PS2}
		\State Calculate objective function values of the new trial solution (Section~\ref{sec:obj}).
		\If{$x^{+}$ is better than the current solution} 
		\State replace the current solution by $x^{+}$
		\EndIf			
		\EndIf
		\If {$x^{+}$ is worse than the current solution} 
		\State $\rho \leftarrow \frac{\rho}{2}$
		\EndIf
		\EndWhile
	\end{algorithmic}
\end{algorithm}

\subsection{Embedding within Pareto-based Techniques}
This subsection presents how to embed the proposed multi-objective JPEG image compression strategy into two well-known Pareto-based techniques, NSGA-II and NSGAIII. 
\subsubsection{EnNSGAII Algorithm}
EnNSGAII algorithm is a Pareto-based technique and generates a set of solutions instead of a single solution. EnNSGAII employs the NGSA-II algorithm for the optimisation process introduced in Section~\ref{sec:nsga2}. We have used the same operators for EnNSGAII algorithm including non-dominated sorting, elite preserving operator, crowding distance, and selection operator. For the evolutionary step, we have used similar operators to the ones used in EnMOGA. In other words, we have used tournament selection, SBX crossover, and Polynomial mutation. Algorithm~\ref{alg:NSGA2} presents the pseudo-code for the EnNSGAII algorithm.
\begin{algorithm}[t!]
	\caption{EnNSGAII algorithm in the form of Pseudo-code.}
	\label{alg:NSGA2}
	\begin{algorithmic}[1]
		\State // $L$/$U$: lower/upper bound;  $N_{pop}$: number of bids; $N_{var}$: number of variables;$NFE_{\max}$:  maximum number of function evaluations; $prob$: the probability of a crossover, $\eta$: the distribution index
		\State \
		\State Initialise population of $N_{pop}$ candidate solution using the representation introduced in Section~\ref{sec:rep}.
		\State Calculate objective function values of all candidate solutions (Section~\ref{sec:obj}).
		\State Assign (level) rank based on Pareto sorting 
		\State $NFE=N_{pop}$
		\State $iter=0$
		\While {$NFE<=NFE_{\max}$}
		\State $iter=iter+1$
		\State Perform Tournament selection (Section~\ref{sec:EnMOGA}).	
		\State Perform SBX mutation (Section~\ref{sec:EnMOGA}).
		\State Perform Polynomial mutation (Section~\ref{sec:EnMOGA}).
		\State Calculate objective function values of all new candidate solutions (Section~\ref{sec:obj}).
	    \State $R_{t} \leftarrow$ Combine parent and offspring population 
	    \State Assign (level) rank based on Pareto sorting
	    \State Generate sets of non-dominated solutions
	    \State Add solutions to next generation starting from the first front to $N_{pop}$ individuals.
	    \State Determine crowding distance
	    \State Select points on the lower front with high crowding distance 
		\State Update $NEF$	
		\EndWhile
	\end{algorithmic}
\end{algorithm}
\subsubsection{EnNSGAIII Algorithm}
EnNSGAIII is similar to EnNSGAII except that it employs reference directions rather than crowding distance. Therefore, all operators we used for EnNSGAIII are similar to EnNSGAII. In other words, EnNSGAIII employs non-dominated sorting, elite preserving operator, tournament selection, SBX operator, and Polynomial mutation for the optimisation process. The EnNSGAIII algorithm in the form of pseudo-code is given in Algorithm~\ref{alg:NSGA3}.

\begin{algorithm}[t!]
	\caption{EnNSGAIII algorithm in the form of Pseudo-code.}
	\label{alg:NSGA3}
	\begin{algorithmic}[1]
		\State // $L$/$U$: lower/upper bound;  $N_{pop}$: number of bids; $N_{var}$: number of variables;$NFE_{\max}$:  maximum number of function evaluations; $prob$: the probability of a crossover, $\eta$: the distribution index
		\State \
		\State Initialise population of $N_{pop}$ candidate solution using the representation introduced in Section~\ref{sec:rep}.
		\State Calculate objective function values of all candidate solutions (Section~\ref{sec:obj}).
		\State Assign (level) rank based on Pareto sorting 
		\State $NFE=N_{pop}$
		\State $iter=0$
		\While {$NFE<=NFE_{\max}$}
		\State $iter=iter+1$
		\State Perform Tournament selection (Section~\ref{sec:EnMOGA}).	
		\State Perform SBX mutation (Section~\ref{sec:EnMOGA}).
		\State Perform Polynomial mutation (Section~\ref{sec:EnMOGA}).
		\State Calculate objective function values of all new candidate solutions (Section~\ref{sec:obj}).
		\State $R_{t} \leftarrow$ Combine parent and offspring population 
		\State Assign (level) rank based on Pareto sorting
		\State Generate sets of non-dominated solutions
		\State Add solutions to next generation starting from the first front to $N_{pop}$ individuals.
		\State Normalise objective function and create reference set
		\State Assign each member to a reference point 
		\State Compute niche count of each reference point
		\State add new members to the new population based on the niche count
		\State Update $NEF$	
		\EndWhile
	\end{algorithmic}
\end{algorithm}  

\section{Experimental Results}
\label{Sec:exp}
\begin{figure}[ht!]
	\centering
	\begin{subfigure}[b]{0.25\linewidth}
		\includegraphics[width=\linewidth] {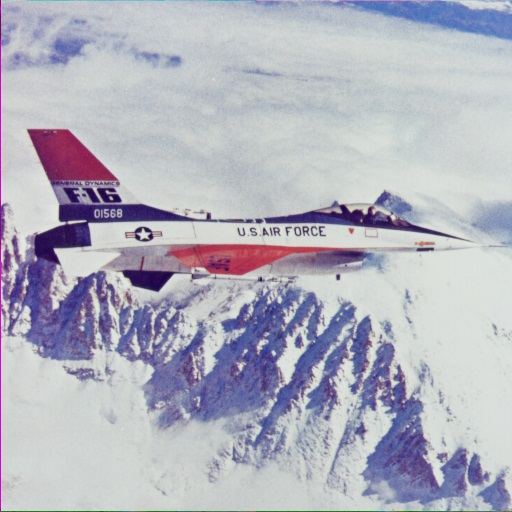}
		\caption{Airplane}
	\end{subfigure} 
    \begin{subfigure}[b]{0.25\linewidth}
    	\includegraphics[width=\linewidth] {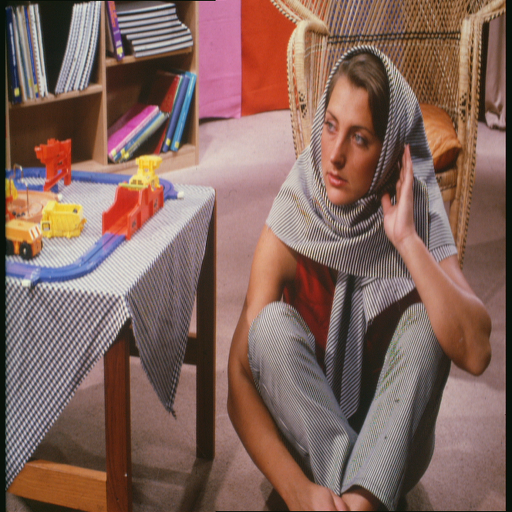}
    	\caption{Barbara}
    \end{subfigure}
	\begin{subfigure}[b]{0.25\linewidth}
		\includegraphics[width=\linewidth] {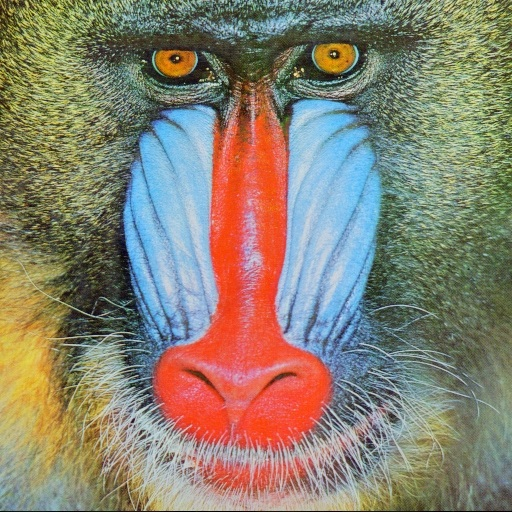}
		\caption{Mandrill}
	\end{subfigure} 
	\begin{subfigure}[b]{0.25\linewidth}
		\includegraphics[width=\linewidth] {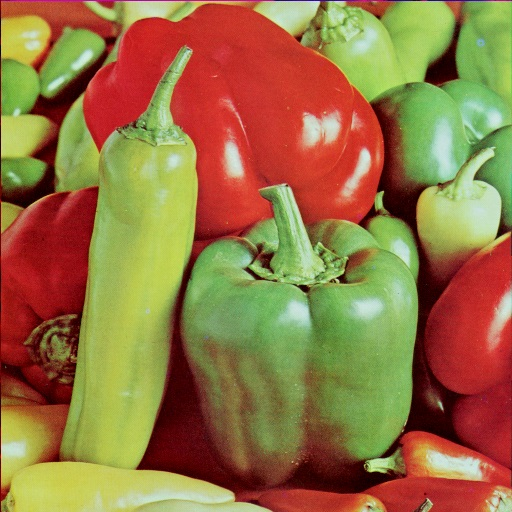}
		\caption{Peppers}
	\end{subfigure}
	\begin{subfigure}[b]{0.25\linewidth}
		\includegraphics[width=\linewidth] {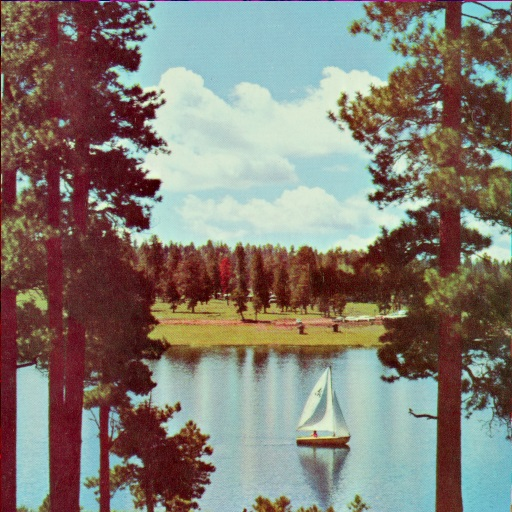}
		\caption{Sailboat}
	\end{subfigure} 
	\begin{subfigure}[b]{0.25\linewidth}
		\includegraphics[width=\linewidth] {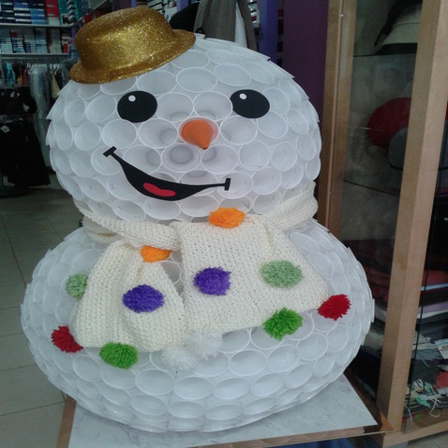}
		\caption{Snowman} 
	\end{subfigure} 
	
	\begin{subfigure}[b]{0.25\linewidth}
		\includegraphics[width=\linewidth] {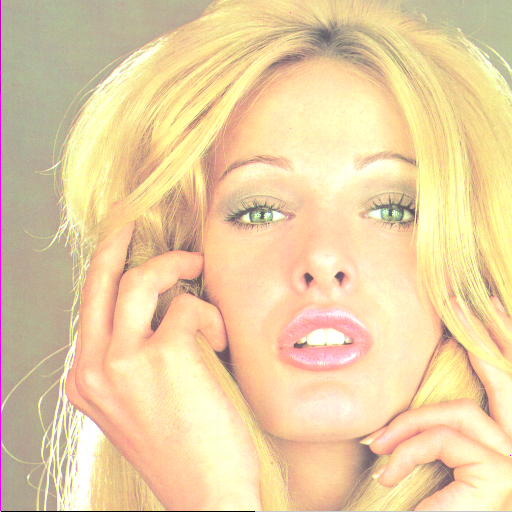}
		\caption{Tiffany}
	\end{subfigure} 
	\begin{subfigure}[b]{0.25\linewidth}
		\includegraphics[width=\linewidth] {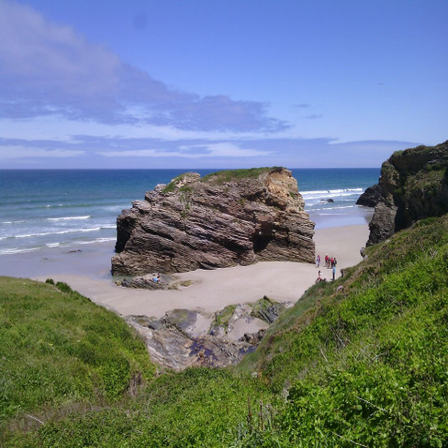}
		\caption{Beach}
	\end{subfigure}
	\begin{subfigure}[b]{0.25\linewidth}
		\includegraphics[width=\linewidth] {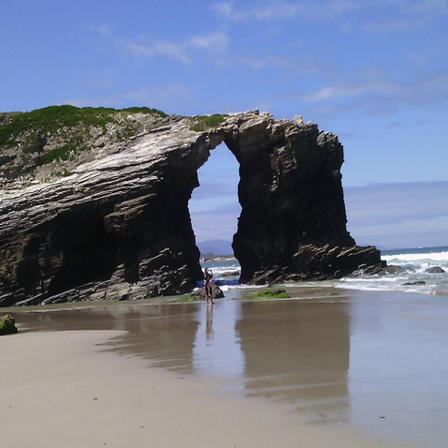}
		\caption{Cathedrals beach}
	\end{subfigure} 
	\begin{subfigure}[b]{0.25\linewidth}
		\includegraphics[width=\linewidth] {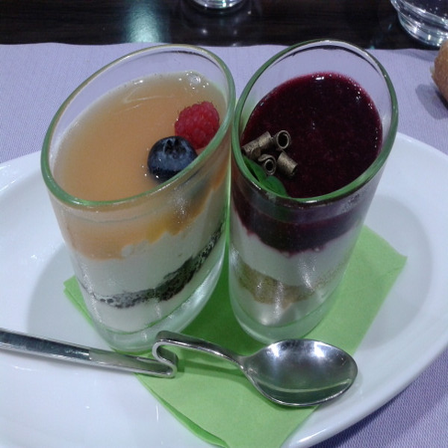}
		\caption{dessert}
	\end{subfigure}
	\begin{subfigure}[b]{0.25\linewidth}
		\includegraphics[width=\linewidth] {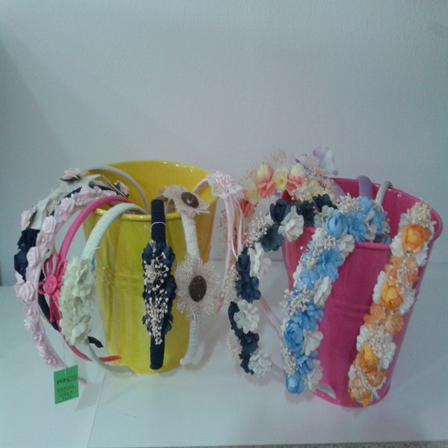}
		\caption{headbands}
	\end{subfigure}
	\begin{subfigure}[b]{0.25\linewidth}
		\includegraphics[width=\linewidth] {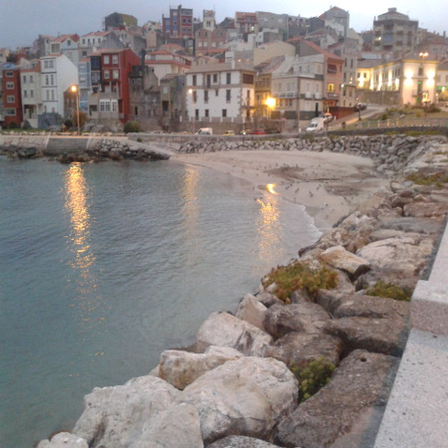}
		\caption{landscape}
	\end{subfigure} 
	
	\caption{Benchmark images.}
	\label{fig:bench}
\end{figure}

To demonstrate the superiority of our proposed strategy, an extensive set of experiments is provided. To this end, we have used 7 popular benchmark images in image compression, including, \textit{Airplane}, \textit{Barbara}, \textit{Lena}, \textit{Mandrill}, \textit{Peppers}, \textit{Tiffany}, and\textit{ Sailboat},  as well as 6 images suggested in~\cite{color_quantisation_SFLA} for image quantisation benchmarking, including, \textit{Snowman}, \textit{Beach}, \textit{Cathedrals beach}, \textit{Dessert}, \textit{Headbands}, and \textit{Landscape}. Figure~\ref{fig:bench} shows the benchmark images. 

Our proposed strategy is embedded in five scalarisation and two Pareto-based methods. All algorithms are run 30 times independently to provide a fair comparison, and their statistical results, including average and standard deviation, are presented. The population size and the number of function evaluations for all algorithms are set to 50 and 1000, respectively. For other parameters, we used the default parameters that can be seen in Table~\ref{tab:parameter}.

\begin{table}[!htbp]
	\centering
	\caption{Parameter settings.}
	\begin{tabular}{l|l|c}
		Algorithm                   & Parameter            & Value \\ \hline
		\multirow{4}{*}{EnMOGA}     & $Prob$ for crossover   & 0.9     \\
		& $\eta$ for crossover & 20     \\
		& $Prob$ for mutation    & 0.3     \\
		& $\eta$ for mutation  & 20     \\ \hline
		EnMOPSO                     & -                   & -     \\ \hline
		EnMODE                      & CR                   & 0.2     \\ \hline
		EnMOES                      & -             & -     \\ \hline
		EnMOPS                      & $\rho$               & 0.5     \\ \hline
		\multirow{4}{*}{EnNSGA-II}  & $Prob$ for crossover   & 0.9     \\
		& $\eta$ for crossover & 20     \\
		& $Prob$ for mutation    & 0.3     \\
		& $\eta$ for mutation  & 20    \\ \hline
		\multirow{4}{*}{EnNSGA-III} & $Prob$ for crossover   & 0.9     \\ 
		& $\eta$ for crossover & 20     \\
		& $Prob$ for mutation    & 0.3     \\
		& $\eta$ for mutation  & 20    \\ \hline
	\end{tabular}
    	\label{tab:parameter}
\end{table}

All algorithms are implemented in Python and with the Pymoo framework~\cite{pymoo}, an open-source framework including state-of-the-art single-and multi-objective algorithms as well as features related to multi-objective optimisation such as visualisation, introduced in 2020.

\subsection{Results of Scalarisation Approaches}
This section aims to find answers to two central questions, as follows:
\begin{compactitem}
\item Is the proposed strategy able to provide higher quality results than the baseline algorithm? (Here, the baseline algorithm means the standard JPEG compression method).
\item Among the embedded scalarisation methods, which algorithm has been able to provide a better solution?	
\end{compactitem}	

To this end, Table~\ref{Tab:scaler} compares the results of the baseline algorithm with other algorithms in terms of mean and standard deviation of the objective function yielded by each algorithm. All algorithms except the baseline algorithm are run 30 times since the baseline algorithm is deterministic. Therefore, we do not provide any standard deviation for the baseline algorithm. Also, the rank of each algorithm per image is indicated from the smallest mean to the highest mean in Table~\ref{Tab:scaler}. In the last row of the table, the average rank of each algorithm and subsequently, the overall ranks are reported as well.

From Table~\ref{Tab:scaler}, and by a comparison between the baseline algorithm with others, we can observe that the baseline algorithm in all cases and in comparison to all algorithms achieves the worst results. For instance, for the \textit{Airplane} image, the objective function for the baseline algorithm is 1.7839, while for others, it is between 1.4185 and 1.4700, indicating a significant improvement in the proposed strategy. Therefore, in short, we can say that our strategy, independent of the embedding algorithm, can provide competitive results compared to the baseline algorithm. 

As mentioned, we employed five scalarisation methods. Here, we compare the results of scalarisation methods together. The results can be seen in Table~\ref{Tab:scaler}. From the table, we can observe that EnMOGA can achieve the first rank in 6 out of 13 images and the second rank in 7 out of 13 images. Also, EnMOPS is placed in the first rank with seven cases, and in the second rank with five images. EnMOES achieved the fifth or worst rank among the embedding algorithms in all cases. Therefore, from the last row of the table, we can say that EnMOGA and EnMOPS provide the best average rank and, subsequently, overall rank, while EnMOES provides the highest average and overall ranks.  
\begin{table}[!htbp]
	\centering
	\caption{A comparison between different scalarisation approaches and the baseline algorithm in terms of objective function.}
	\label{Tab:scaler}
	\begin{tabular}{l|l|c|ccccc}
		Images             &             & Baseline & EnMOGA & EnMOPSO & EnMODE & EnMOES & EnMOPS \\ \hline
		Airplane           & Mean        & 1.7839   & 1.4227  & 1.4370   & 1.4596  & 1.4700  & 1.4185   \\
		& Std.        & -        & 0.0055  & 0.0064   & 0.0051  & 0.0034  & 0.0159   \\
		& rank        & 6        & 2       & 3        & 4       & 5       & 1        \\ \hline
		Barbara            & Mean        & 1.7647   & 1.3571  & 1.3688   & 1.4052  & 1.4224  & 1.3731   \\
		& Std.        & -        & 0.0076  & 0.0079   & 0.0079  & 0.0060  & 0.0265   \\
		& rank        & 6        & 1       & 2        & 4       & 5       & 3        \\ \hline
		Lena               & Mean        & 1.8552   & 1.4312  & 1.4475   & 1.4665  & 1.4794  & 1.4337   \\
		& Std.        & -        & 0.0062  & 0.0072   & 0.0054  & 0.0029  & 0.0205   \\
		& rank        & 6        & 1       & 3        & 4       & 5       & 2        \\ \hline
		Mandrill           & Mean        & 1.8961   & 1.6098  & 1.6329   & 1.6666  & 1.6767  & 1.5703   \\
		& Std.        & -        & 0.0073  & 0.0084   & 0.0041  & 0.0049  & 0.0196   \\
		& rank        & 6        & 2       & 3        & 4       & 5       & 1        \\\hline
		Peppers            & Mean        & 1.8604   & 1.5168  & 1.5313   & 1.5409  & 1.5507  & 1.5004   \\
		& Std.        & -        & 0.0046  & 0.0048   & 0.0044  & 0.0036  & 0.0120   \\
		& rank        & 6        & 2       & 3        & 4       & 5       & 1        \\ \hline
		Sailboat           & Mean        & 1.8458   & 1.5670  & 1.5839   & 1.6015  & 1.6103  & 1.5379   \\
		& Std.        & -        & 0.0048  & 0.0063   & 0.0033  & 0.0027  & 0.0150   \\
		& rank        & 6        & 2       & 3        & 4       & 5       & 1        \\ \hline
		Snowman            & Mean        & 1.8239   & 1.4041  & 1.4150   & 1.4467  & 1.4611  & 1.4120   \\
		& Std.        & -        & 0.0071  & 0.0075   & 0.0080  & 0.0054  & 0.0189   \\
		& rank        & 6        & 1       & 3        & 4       & 5       & 2        \\ \hline
		Tiffany           & Mean        & 1.8027   & 1.5109  & 1.5147   & 1.5229  & 1.5282  & 1.4867   \\
		& Std.        & -        & 0.0018  & 0.0024   & 0.0026  & 0.0023  & 0.0093   \\
		& rank        & 6        & 2       & 3        & 4       & 5       & 1        \\ \hline
		Beach              & Mean        & 1.8047   & 1.4847  & 1.4956   & 1.5282  & 1.5386  & 1.4723   \\
		& Std.        & -        & 0.0061  & 0.0081   & 0.0058  & 0.0043  & 0.0186   \\
		& rank        & 6        & 2       & 3        & 4       & 5       & 1        \\ \hline
		Cathedrals beach         & Mean        & 1.7551   & 1.3648  & 1.3748   & 1.4013  & 1.4110  & 1.3647   \\
		& Std.        & -        & 0.0076  & 0.0080   & 0.0056  & 0.0056  & 0.0175   \\
		& rank        & 6        & 2       & 3        & 4       & 5       & 1        \\ \hline
		Dessert            & Mean        & 1.8195   & 1.3760  & 1.3859   & 1.4113  & 1.4220  & 1.3804   \\
		& Std.        & -        & 0.0067  & 0.0066   & 0.0076  & 0.0053  & 0.0180   \\
		& rank        & 6        & 1       & 3        & 4       & 5       & 2        \\ \hline
		Headbands          & Mean        & 1.8027   & 1.3879  & 1.3971   & 1.4201  & 1.4269  & 1.3901   \\
		& Std.        & -        & 0.0047  & 0.0058   & 0.0042  & 0.0037  & 0.0141   \\
		& rank        & 6        & 1       & 3        & 4       & 5       & 2        \\ \hline
		Landscape          & Mean        & 1.8456   & 1.4272  & 1.4386   & 1.4731  & 1.4940  & 1.4303   \\
		& Std.        & -        & 0.0066  & 0.0081   & 0.0087  & 0.0058  & 0.0203   \\
		& rank        & 6        & 1       & 3        & 4       & 5       & 2        \\ \hline
		\multicolumn{2}{l}{Average rank} & 6        & 1.54    & 2.92     & 4.00    & 5.00    & 1.54     \\ \hline
		\multicolumn{2}{l}{Overall rank} & 6        & 1.5     & 3        & 4       & 5       & 1.5 \\ \hline    
	\end{tabular}
\end{table}

Due to the non-deterministic behaviour of metaheuristic algorithms, non-parametric statistical is obligatory. In this case, the alternative hypothesis $H_{1}$ denotes a statistically significant difference between the algorithms, while the null hypothesis $H_{0}$ states that there is no statistical difference between the two algorithms. The null hypothesis is the initial statistical assertion, and the alternative hypothesis would be accepted if the null hypothesis were to be shown to be false. To this end, we carried out Wilcoxon signed rank test~\cite{tutorial_statistical} at 5\% significance level based on the mean objective function value to compare the results statistically.

Table~\ref{tab:wilc} shows that EnMOGA and EnMOPS perform statistically superior to other algorithms since both win in 4 cases. Also, EnMOPS and EnMOGA are statistically the same. The overall following best working algorithm is EnMOPSO (3 wins, 2 losses). Among the scalarisation algorithms, EnMOES performs worst (1 win, 4 losses). Again, the baseline algorithm (0 win, 5 losses) fails against all proposed scalarisation methods. 

\begin{table}[!htbp]
	\caption{Results of Wilcoxon signed rank test based on mean objective function value. $\ddagger$, $\dagger$ , and $\approx$ indicate that the algorithm in the corresponding row is statistically better than, worse than, or similar to the algorithm in the corresponding column. The last column summarises the algorithms' total wins (w), ties (t), and losses (l).}
	\label{tab:wilc}
	\begin{tabular}{l|cccccc|c}
		& Baseline & EnMOGA & EnMOPSO & EnMODE & EnMOES & EnMOPS & w/t/l \\ \hline \hline
		Baseline & \cellcolor{gray!5}  & $\dagger$       & $\dagger$        & $\dagger$       & $\dagger$       & $\dagger$        & 0/0/5 \\
		EnMOGA & $\ddagger$        &  \cellcolor{gray!5}  & $\ddagger$        & $\ddagger$       & $\ddagger$       & $\approx$        & 4/1/0 \\
		EnMOPSO & $\ddagger$        & $\dagger$       & \cellcolor{gray!5}  & $\ddagger$       & $\ddagger$       & $\dagger$        & 3/0/2 \\
		EnMODE  & $\ddagger$        & $\dagger$       & $\dagger$        & \cellcolor{gray!5}  & $\ddagger$       & -        & 2/0/3 \\
		EnMOES  & $\ddagger$        & $\dagger$       & $\dagger$        & $\dagger$       & \cellcolor{gray!5} & $\dagger$        & 1/0/4 \\
		EnMOPS & $\ddagger$        & $\approx$       & $\ddagger$        & $\ddagger$       & $\ddagger$       & \cellcolor{gray!5} & 4/1/0
	\end{tabular}
\end{table}

\begin{figure}[!htbp]
	\centering
	\begin{subfigure}[b]{0.32\linewidth}
		\includegraphics[width=\linewidth] {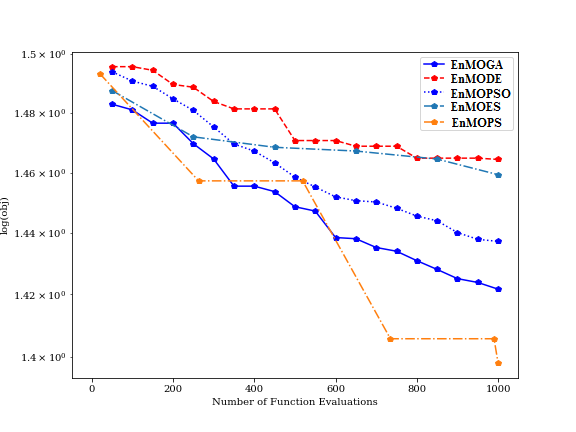}
		\caption{Airplane}
	\end{subfigure} 
	\begin{subfigure}[b]{0.32\linewidth}
		\includegraphics[width=\linewidth] {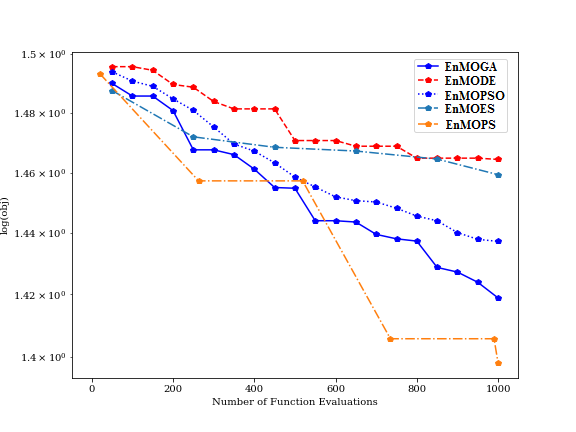}
		\caption{Barbara}
	\end{subfigure}
	\begin{subfigure}[b]{0.32\linewidth}
		\includegraphics[width=\linewidth] {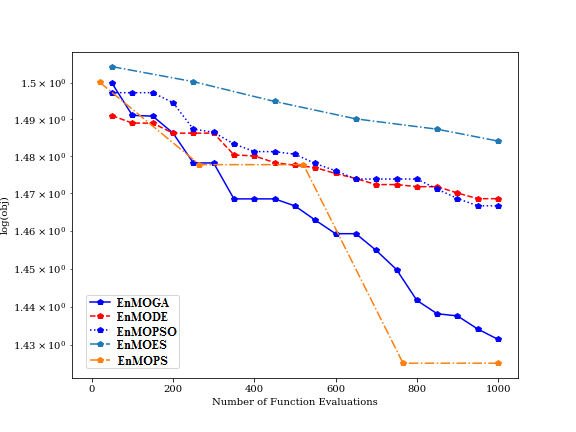}
		\caption{Lena}
	\end{subfigure} 
	\begin{subfigure}[b]{0.32\linewidth}
		\includegraphics[width=\linewidth] {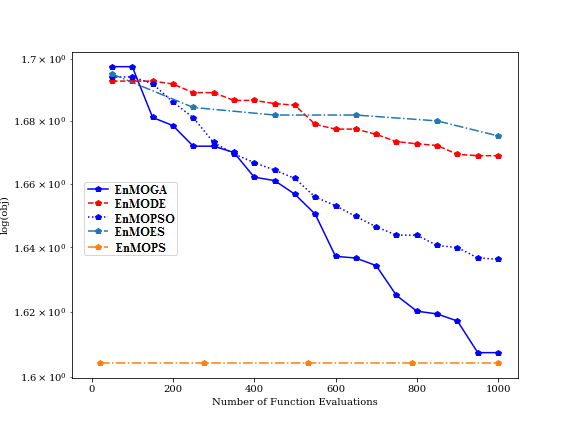}
		\caption{Mandrill}
	\end{subfigure}
	\begin{subfigure}[b]{0.32\linewidth}
		\includegraphics[width=\linewidth] {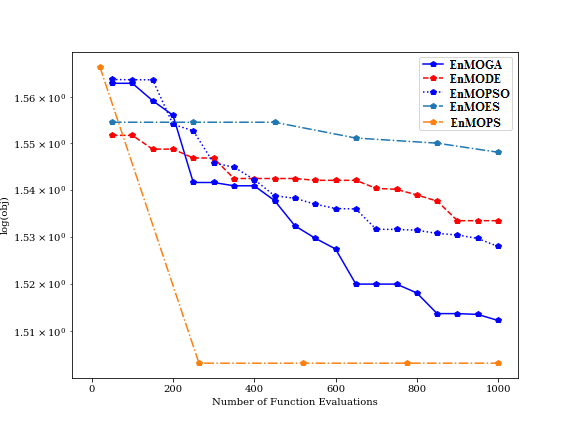}
		\caption{Peppers}
	\end{subfigure} 
	\begin{subfigure}[b]{0.32\linewidth}
		\includegraphics[width=\linewidth] {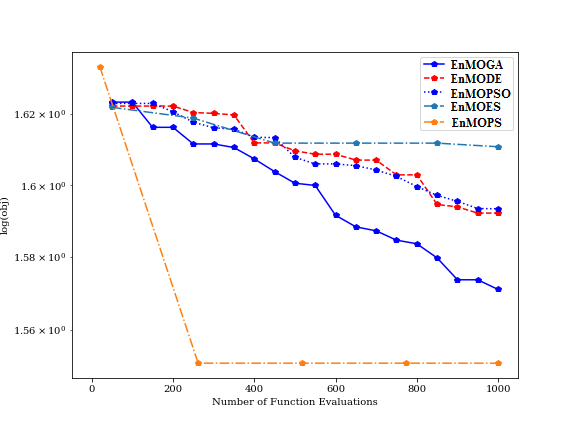}
		\caption{Sailboat} 
	\end{subfigure} 
	
	\begin{subfigure}[b]{0.32\linewidth}
		\includegraphics[width=\linewidth] {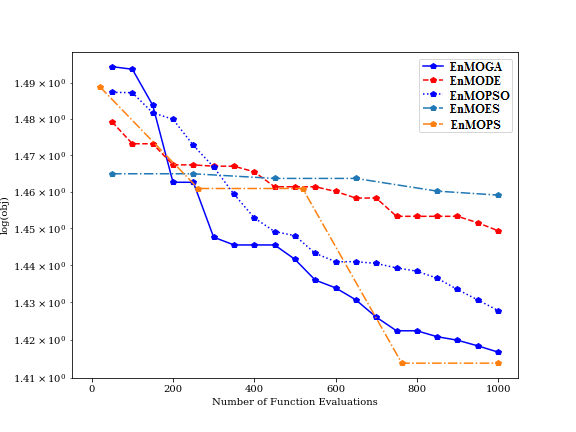}
		\caption{Snowman}
	\end{subfigure} 
	\begin{subfigure}[b]{0.32\linewidth}
		\includegraphics[width=\linewidth] {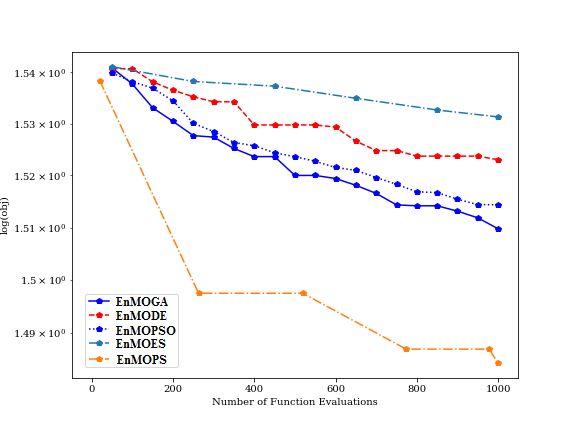}
		\caption{tiffany}
	\end{subfigure}
	\begin{subfigure}[b]{0.32\linewidth}
		\includegraphics[width=\linewidth] {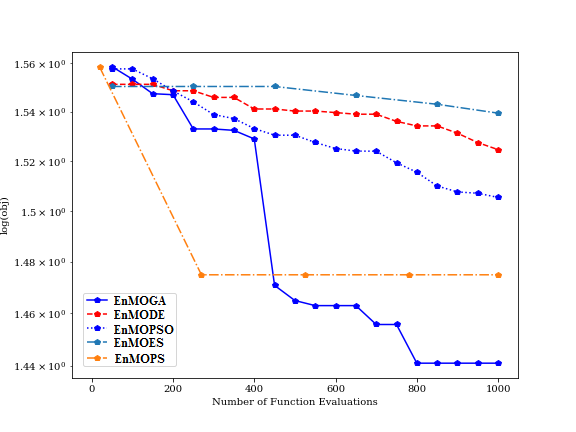}
		\caption{Beach}
	\end{subfigure} 
	\begin{subfigure}[b]{0.32\linewidth}
		\includegraphics[width=\linewidth] {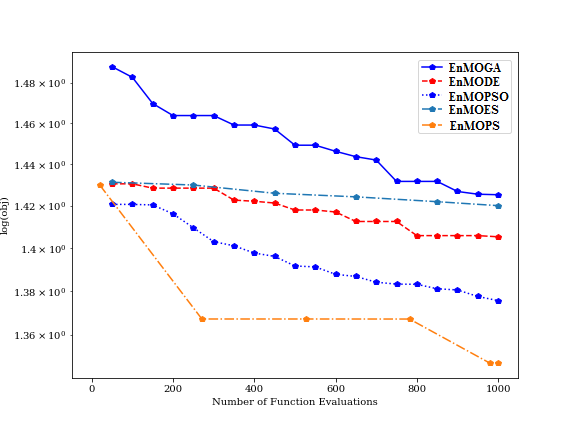}
		\caption{Cathedrals beach}
	\end{subfigure}
	\begin{subfigure}[b]{0.32\linewidth}
		\includegraphics[width=\linewidth] {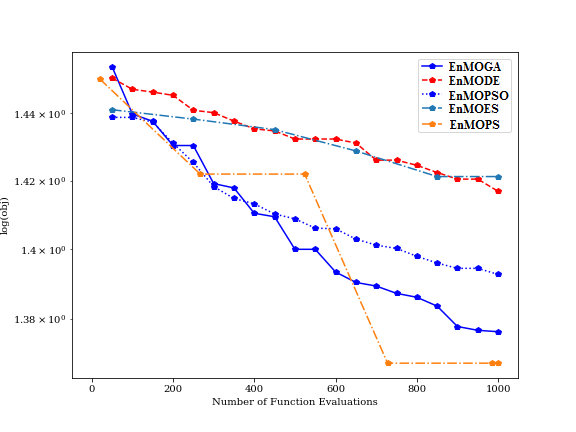}
		\caption{Dessert}
	\end{subfigure}
	\begin{subfigure}[b]{0.32\linewidth}
		\includegraphics[width=\linewidth] {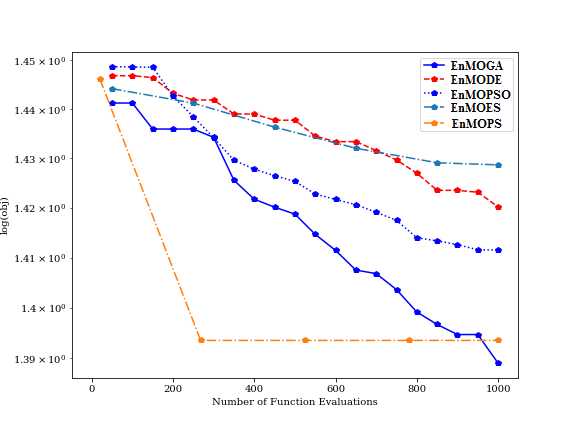}
		\caption{Headbands}
	\end{subfigure} 
	
	\begin{subfigure}[b]{0.32\linewidth}
		\includegraphics[width=\linewidth] {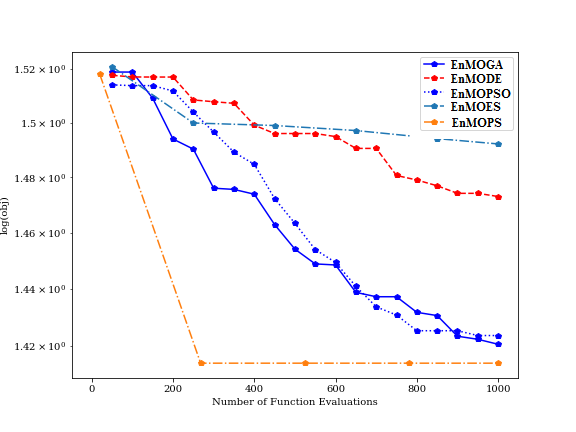}
		\caption{Landscape}
	\end{subfigure}

	\caption{Convergence curves for scalarisation approaches.}
	\label{Con_scal}
\end{figure}

We also investigated the convergence curves for all algorithms. Figure~\ref{Con_scal} shows plots of objective function values against the number of function evaluations on all images and for a single random run. It can be seen that EnMOPS and EnMOGA have faster convergence compared to other algorithms, while EnMOES suffers from low-speed convergence.

\subsection{Results of Pareto-based Approaches}

The evaluation method of the Pareto-based approaches differs from the scalarisation approaches because the Pareto-based approaches result in several solutions while the scalarisation methods produce only one solution. There are several measures to validate the results of Pareto-based approaches, such as Generational Distance (GD)~\cite{GD} and Inverted Generational Distance (IGD)~\cite{IGD}, while here, we cannot use these because they require the true Pareto front, which is not available in this problem. Therefore, we used an alternative measure, called hyper-volume (HV)~\cite{hyper_volume}, which does not require a true Pareto-front to validate the results. The HV measure is regarded as a fair measure among other criteria~\cite{hyper_volume2} so that HV can take into consideration both closeness to the optimal solution and being well-distributed along the whole Pareto front. HV measure determines the area/volume that, in relation to a reference point, is dominated by the given set of solutions. A higher value of HV measure in a minimisation problem shows a better quality of the solution. 

This paper integrated the proposed strategy into two Pareto-based algorithms, including NSGA-II and NSGA-III. The results based on the HV measure are given in Table~\ref{tab:hv}. The table shows that EnNSGAII outperforms EnNSGAIII in 10 out of 13 cases, while it fails in 3 cases. From the last row of the table, we can observe that the average rank of EnNSGAII is lower than EnNSGAIII. In other words, EnNSGAII overcomes EnNSGAIII.

To perform a deeper analysis, we also conducted a Wilcoxon signed rank test on the results. The achieved $p$-value is 0.0574, which means that there is a statistical difference between the two algorithms only at a 10\% significance level.

\begin{table}[!htbp]
	\centering
	\caption{A comparison between EnNSGAII and EnNSGAIII in terms of the HV measure. }
	\label{tab:hv}
	\begin{tabular}{l|l|cc}
		Images       &      & EnNSGAII & EnNSGAIII \\ \hline
		Airplane     & Mean & 7.0803         & 7.0920          \\
		& Std. & 0.1805         & 0.2133          \\
		& rank & 2              & 1               \\ \hline
		Barbara      & Mean & 7.1164         & 7.0369          \\
		& Std. & 0.1717         & 0.2482          \\
		& rank & 1              & 2               \\ \hline
		Lena         & Mean & 7.1206         & 7.1436          \\
		& Std. & 0.2085         & 0.1077          \\
		& rank & 2              & 1               \\ \hline
		Mandrill     & Mean & 7.0618         & 7.0585          \\
		& Std. & 0.0483         & 0.0646          \\
		& rank & 1              & 2               \\ \hline
		Peppers      & Mean & 7.2614         & 6.6218          \\
		& Std. & 0.0634       & 0.5518        \\
		& rank & 1              & 2               \\ \hline
		Sailboat     & Mean & 7.0653       & 7.0424        \\
		& Std. & 0.1971       & 0.0986        \\
		& rank & 1              & 2               \\ \hline
		Snowman      & Mean & 7.1900       & 7.0897        \\
		& Std. & 0.0599       & 0.0727        \\
		& rank & 1              & 2               \\ \hline
		Tiffany      & Mean & 6.9587       & 7.1036        \\
		& Std. & 0.4292       & 0.1055        \\
		& rank & 2              & 1               \\ \hline
		Beach        & Mean & 7.078745       & 6.960397        \\
		& Std. & 0.03692      & 0.1994       \\
		& rank & 1              & 2               \\ \hline
		Cathedrals beach   & Mean & 7.1510       & 7.1149        \\
		& Std. & 0.1207      & 0.0571        \\
		& rank & 1              & 2               \\ \hline
		Dessert      & Mean & 7.1242       & 7.0103         \\
		& Std. & 0.2396       & 0.3402       \\
		& rank & 1              & 2               \\ \hline
		Heatbands    & Mean & 7.2309       & 7.1754        \\
		& Std. & 0.0521       & 0.1084       \\
		& rank & 1              & 2               \\ \hline
		Landscape    & Mean & 7.1799       & 7.1417        \\
		& Std. & 0.0765       & 0.0837       \\
		& rank & 1              & 2               \\ \hline
		Average rank &      & 1.23           & 1.77            \\ \hline
		Overall rank &      & 1              & 2    \\ \hline          
	\end{tabular}
\end{table}
\begin{figure}[!htbp]
	\centering
	\begin{subfigure}[b]{0.32\linewidth}
		\includegraphics[width=\linewidth] {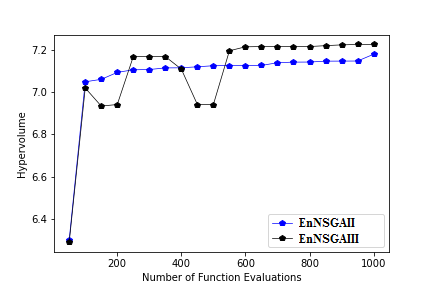}
		\caption{Airplane}
	\end{subfigure} 
	\begin{subfigure}[b]{0.32\linewidth}
		\includegraphics[width=\linewidth] {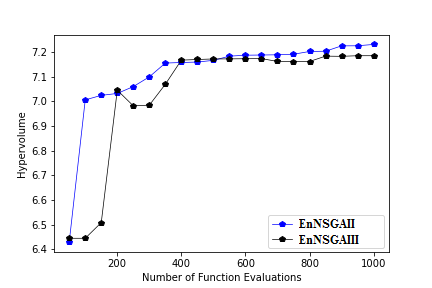}
		\caption{Barbara}
	\end{subfigure}
	\begin{subfigure}[b]{0.32\linewidth}
		\includegraphics[width=\linewidth] {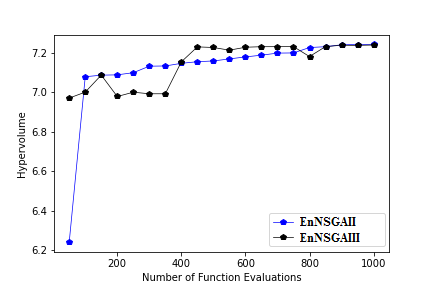}
		\caption{Lena}
	\end{subfigure} 
	\begin{subfigure}[b]{0.32\linewidth}
		\includegraphics[width=\linewidth] {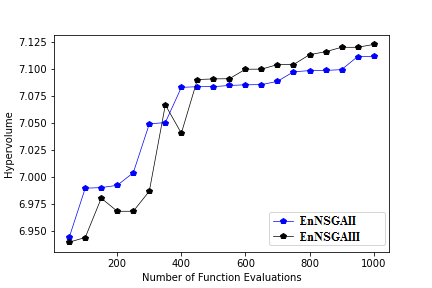}
		\caption{Mandrill}
	\end{subfigure}
	\begin{subfigure}[b]{0.32\linewidth}
		\includegraphics[width=\linewidth] {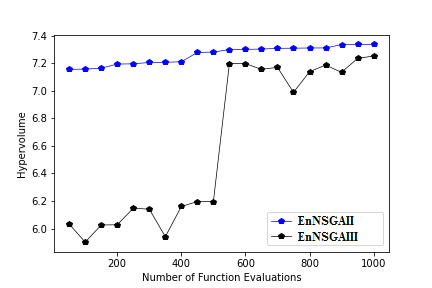}
		\caption{Peppers}
	\end{subfigure} 
	\begin{subfigure}[b]{0.32\linewidth}
		\includegraphics[width=\linewidth] {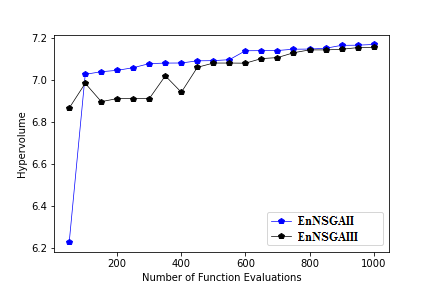}
		\caption{Sailboat} 
	\end{subfigure} 
	
	\begin{subfigure}[b]{0.32\linewidth}
		\includegraphics[width=\linewidth] {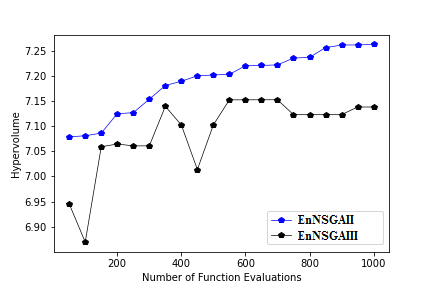}
		\caption{Snowman}
	\end{subfigure} 
	\begin{subfigure}[b]{0.32\linewidth}
		\includegraphics[width=\linewidth] {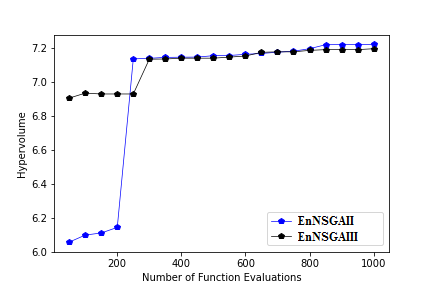}
		\caption{tiffany}
	\end{subfigure}
	\begin{subfigure}[b]{0.32\linewidth}
		\includegraphics[width=\linewidth] {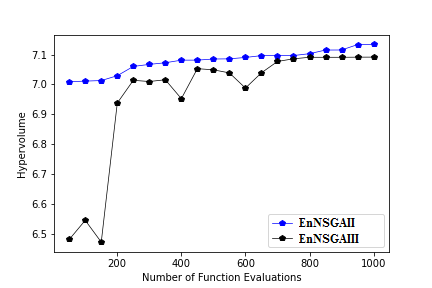}
		\caption{Beach}
	\end{subfigure} 
	\begin{subfigure}[b]{0.32\linewidth}
		\includegraphics[width=\linewidth] {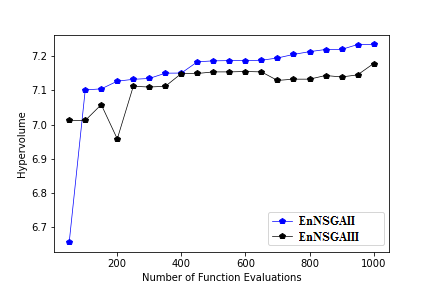}
		\caption{Cathedrals beach}
	\end{subfigure}
	\begin{subfigure}[b]{0.32\linewidth}
		\includegraphics[width=\linewidth] {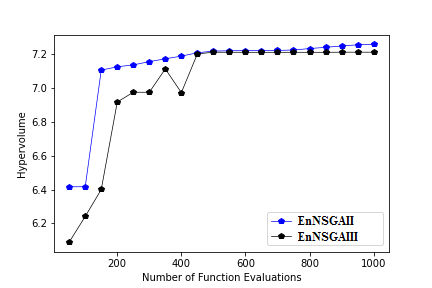}
		\caption{Dessert}
	\end{subfigure}
	\begin{subfigure}[b]{0.32\linewidth}
		\includegraphics[width=\linewidth] {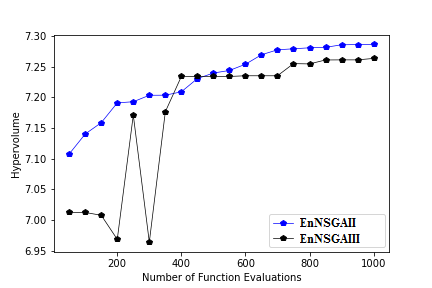}
		\caption{Headbands}
	\end{subfigure} 
	
	\begin{subfigure}[b]{0.32\linewidth}
		\includegraphics[width=\linewidth] {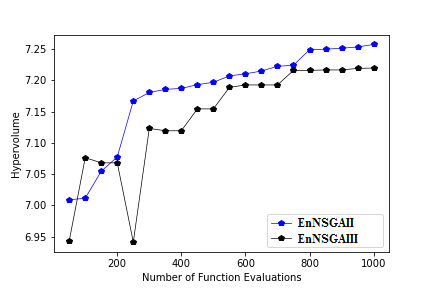}
		\caption{Landscape}
	\end{subfigure}
	
	\caption{Convergence curves for Pareto-based approaches in terms of the HV measure.}
	\label{Con_pareto}
\end{figure}

\begin{figure}[!htbp]
	\centering
	\begin{subfigure}[b]{0.32\linewidth}
		\includegraphics[width=\linewidth] {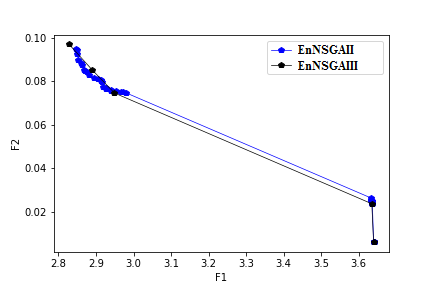}
		\caption{Airplane}
	\end{subfigure} 
	\begin{subfigure}[b]{0.32\linewidth}
		\includegraphics[width=\linewidth] {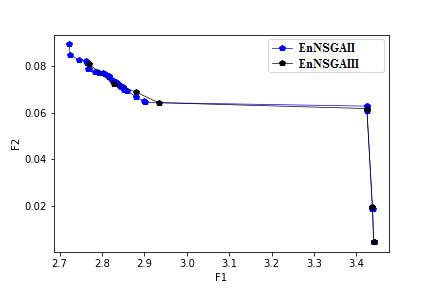}
		\caption{Barbara}
	\end{subfigure}
	\begin{subfigure}[b]{0.32\linewidth}
		\includegraphics[width=\linewidth] {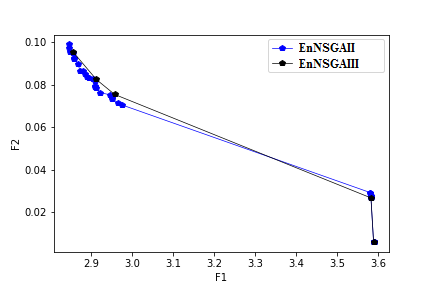}
		\caption{Lena}
	\end{subfigure} 
	\begin{subfigure}[b]{0.32\linewidth}
		\includegraphics[width=\linewidth] {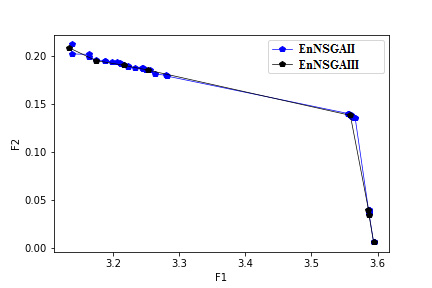}
		\caption{Mandrill}
	\end{subfigure}
	\begin{subfigure}[b]{0.32\linewidth}
		\includegraphics[width=\linewidth] {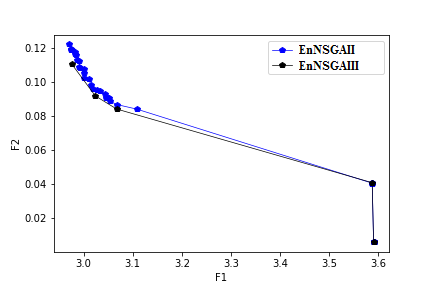}
		\caption{Peppers}
	\end{subfigure} 
	\begin{subfigure}[b]{0.32\linewidth}
		\includegraphics[width=\linewidth] {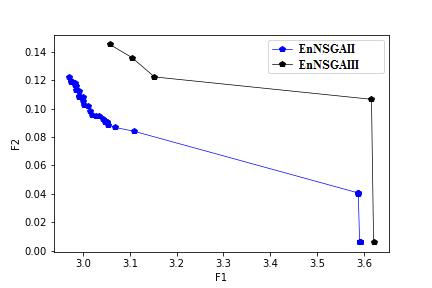}
		\caption{Sailboat} 
	\end{subfigure} 
	
	\begin{subfigure}[b]{0.32\linewidth}
		\includegraphics[width=\linewidth] {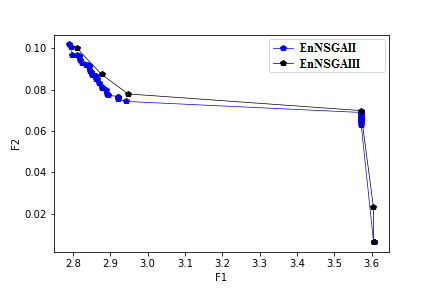}
		\caption{Snowman}
	\end{subfigure} 
	\begin{subfigure}[b]{0.32\linewidth}
		\includegraphics[width=\linewidth] {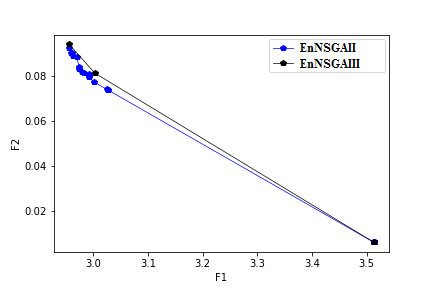}
		\caption{tiffany}
	\end{subfigure}
	\begin{subfigure}[b]{0.32\linewidth}
		\includegraphics[width=\linewidth] {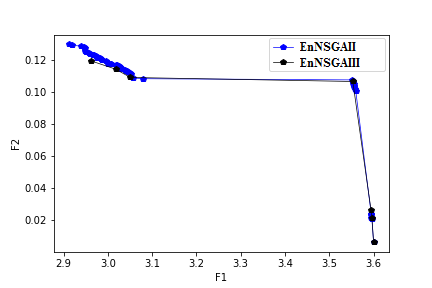}
		\caption{Beach}
	\end{subfigure} 
	\begin{subfigure}[b]{0.32\linewidth}
		\includegraphics[width=\linewidth] {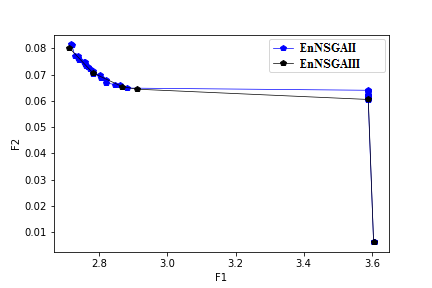}
		\caption{Cathedrals beach}
	\end{subfigure}
	\begin{subfigure}[b]{0.32\linewidth}
		\includegraphics[width=\linewidth] {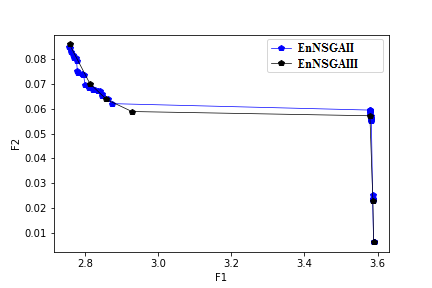}
		\caption{Dessert}
	\end{subfigure}
	\begin{subfigure}[b]{0.32\linewidth}
		\includegraphics[width=\linewidth] {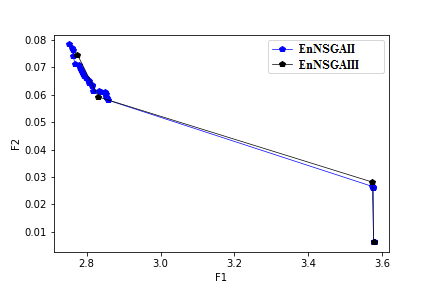}
		\caption{Headbands}
	\end{subfigure} 
	
	\begin{subfigure}[b]{0.32\linewidth}
		\includegraphics[width=\linewidth] {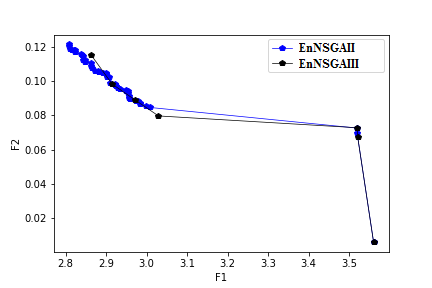}
		\caption{Landscape}
	\end{subfigure}
	
	\caption{A comparison of the Pareto front of the proposed methods for each image.}
	\label{pareto_front}
\end{figure}

Finally, to have a more comprehensive view of the generated Pareto fronts, we plot the Pareto front for the algorithms in Figure~\ref{pareto_front}. It can be seen that EnNSGAII provides more points in the Pareto front compared to EnNSGAIII. 

Since there are two conflicting objective functions, we can not plot convergence curves in terms of objective functions. Therefore, in the next experiment, we indicated the convergence curves in terms of the HV measure rather than objective functions. Figure~\ref{Con_pareto} shows the convergence curves for our two proposed algorithms. It is clear that, in most cases, EnNSGAII provides a faster convergence rate.
 
\subsection{Comparison between Scalarisation and Pareto-based Methods}
Generally speaking, a comparison between scalarisation and Pareto-based methods is not possible since scalarisation methods only generate one solution based on a given set of weights, while Pareto-based methods generate a set of solutions. To tackle this problem, we select one solution from a Pareto-based method as 
		\begin{equation}
\label{Eq:comparison}
PF_{selected}=\min_{i=1}^{N_{PF}} (w_{1}f_{1}^{i}(x)+w_{2}f_{2}^{i}(x)+...+w_{M}f_{M}^{i}(x))
\end{equation}
where $N_{PF}$ is the number of solutions in the generated Pareto front, $w_{1},...,w_{M}$ are the corresponding weights in the scalarisation method, and $f_{1}^{i},...,f_{M}^{i}$ are the objective function values for the $i$-th solution in the Pareto front.

Table~\ref{tab:comparison} shows the results. It can be seen that EnNSGAII can not work better than two others when we select only one solution, as expected since the Pareto-based approaches focus on a set of solutions and not only one solution. Despite the performance of the scalarisation method compared to the Pareto-based  algorithm, we cannot say that the Pareto-based algorithm did not work well because the output of the Pareto-based algorithm is a set of solutions with different weights, while the scalarisation method does not have such an ability. Also, from this experiment, it is worthwhile to mention that if we know the weights of each objective, scalarisation methods are preferable. 

\begin{table}[!htbp]
	\centering
	\caption{A comparison between the scalarisation and Pareto-based methods. The values signify the objective function value defined for scalarisation method and $PF_{selected}$ for EnNSGAII. The best result for a given image is boldfaced. }
	\label{tab:comparison}
	\begin{tabular}{l|ccc}
		\hline
		Images     & EnMOGA & EnMOPS & EnNSGAII   \\ \hline
		Airplane   & 1.4227  & \textbf{1.4185}   & 1.4705 \\ 
		Barbara    & \textbf{1.3571}  & 1.3731   & 1.4048 \\ 
		Lena       & \textbf{1.4312}  & 1.4337   & 1.4720 \\ 
		Mandrill   & 1.6098  & \textbf{1.5703}   & 1.6695 \\ 
		Peppers    & 1.5168  & \textbf{1.5004}   & 1.5462 \\ 
		Sailboat   & 1.5670  & \textbf{1.5379}   & 1.6076 \\ 
		Snowman    & \textbf{1.4041}  & 1.4120   & 1.5212 \\ 
		Tiffany   & 1.5109  & 1.4867   & \textbf{1.4207} \\ 
		Beach      & 1.4847  & 1.4723   & \textbf{1.3987} \\ 
		Cathedrals beach & 1.3648  & \textbf{1.3647}   & 1.4155 \\ 
		Dessert    & \textbf{1.3760}  & 1.3804   & 1.4459 \\ 
		Headbands  & \textbf{1.3879}  & 1.3901   & 1.5244 \\ 
		Landscape  & \textbf{1.4272}  & 1.4303   & 1.4648 \\ \hline
	\end{tabular}
\end{table}
\subsection{Sensitivity Analysis}
The sensitivity analysis of the suggested algorithm's control parameters is examined below. To serve as representatives, we chose two images, namely \textit{Airplane} and \textit{Barbara}. We selected EnMOGA as a scalarisation method, and EnNSGAII as a Pareto-based method, which show better performance, for our experiments.
\subsubsection{Sensitivity to Population Size}

Population size is one of the most critical parameters in metaheuristic algorithms. Metaheuristic algorithms with a large population size usually provide better results than small population size since a large population size supports higher diversity for the population, leading to higher exploration ability due to the recombination of its diverse members~\cite{MicroEvolution01,MicroEvolution02}. Nevertheless, sometimes it is more effective to use a small population size. The term micro-algorithm, $\mu$-algorithm, refers to a metaheuristic algorithm with a small population size~\cite{MicroEvolution02}.

\begin{figure}[!htbp]
	\centering
	\begin{subfigure}[b]{0.8\linewidth}
		\includegraphics[width=\linewidth] {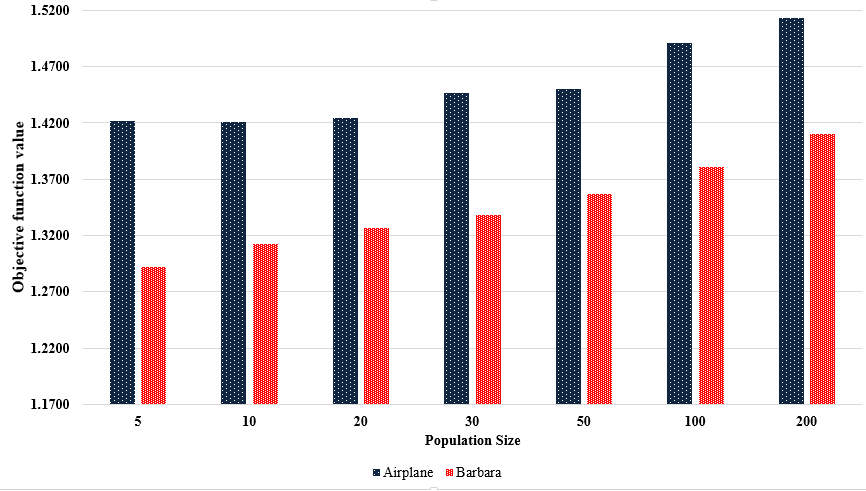}
		\caption{EnMOGA algorithm}
		\label{fig:Obj-GA}
	\end{subfigure} 
	\begin{subfigure}[b]{0.8\linewidth}
		\label{fig:Obj-NSGA2}
		\includegraphics[width=\linewidth] {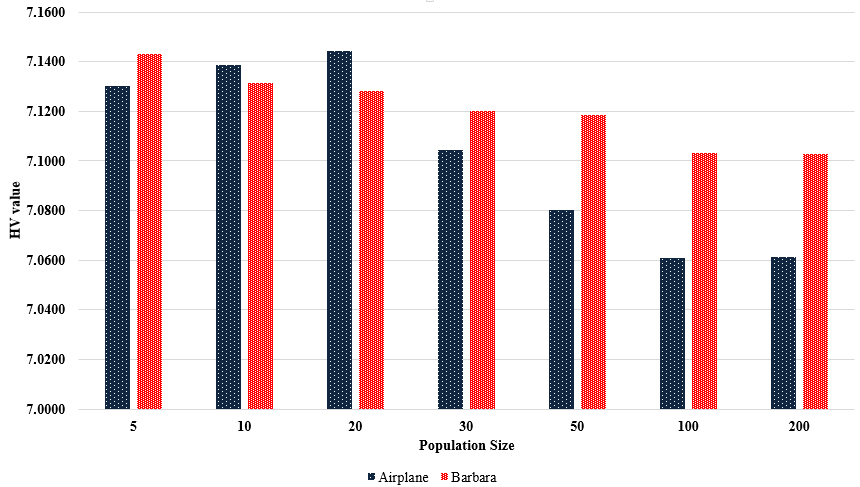}
		\caption{EnNSGAII algorithm}
	\end{subfigure}
	
	\caption{The objective function values obtained with different population sizes.}
	\label{fig:Obj}
\end{figure}

This section aims to investigate the effect of population size on performance. To this end, the population size is set to 5, 10, 20, 30, 50, 100, and 200, while the number of function evaluations is fixed for all algorithms. In other words, for smaller population size, the number of iterations is higher than for larger population size. Figure\ref{fig:Obj}~\subref{fig:Obj-GA} shows the objective function value achieved by different population sizes and for the EnMOGA algorithm. Both images show an upward trend; in other words, a larger population size leads to a higher objective function value. It means that lower population size is preferable. 

The same experiment is performed by EnNSGAII, and the results are given in Figure~\ref{fig:Obj}~\subref{fig:Obj-NSGA2}. For the \textit{Airplane} image, by increasing the population size from 5 to 20, the HV value is also increased, while there is a downward trend by increasing the population size from 20 to 200. For the \textit{Barbara} image, the conditions are a bit different, and the trend is downward for all population sizes. 

In short, it can be said that smaller population sizes can lead to better results for the scalarisation approach, whereas the Pareto-based approach works better with larger population sizes.

\subsubsection{Sensitivity to \textit{prob} and $\eta$ in the Crossover Operator}
\begin{figure}[!htbp]
	\centering
	\begin{subfigure}[b]{0.8\linewidth}
		\includegraphics[width=\linewidth] {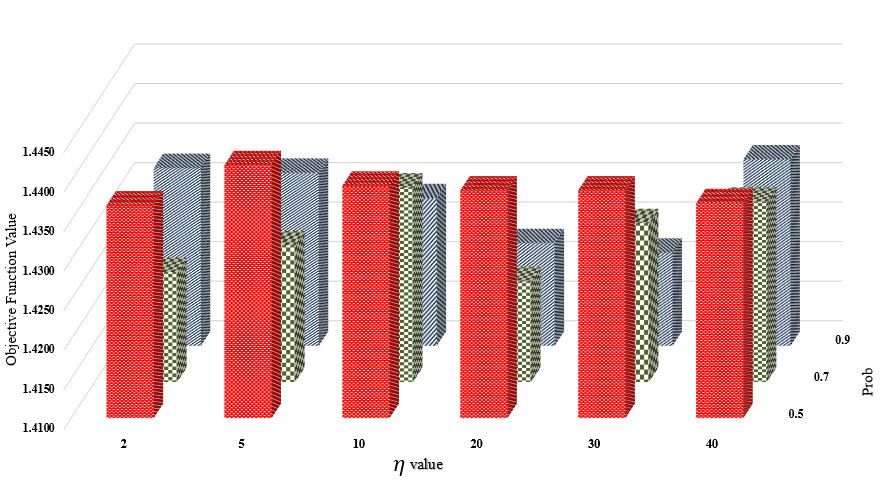}
		\caption{\textit{Airplane} image}
		\label{fig:GA_Crossover_Airplane}
	\end{subfigure} 
	\begin{subfigure}[b]{0.8\linewidth}
		\includegraphics[width=\linewidth] {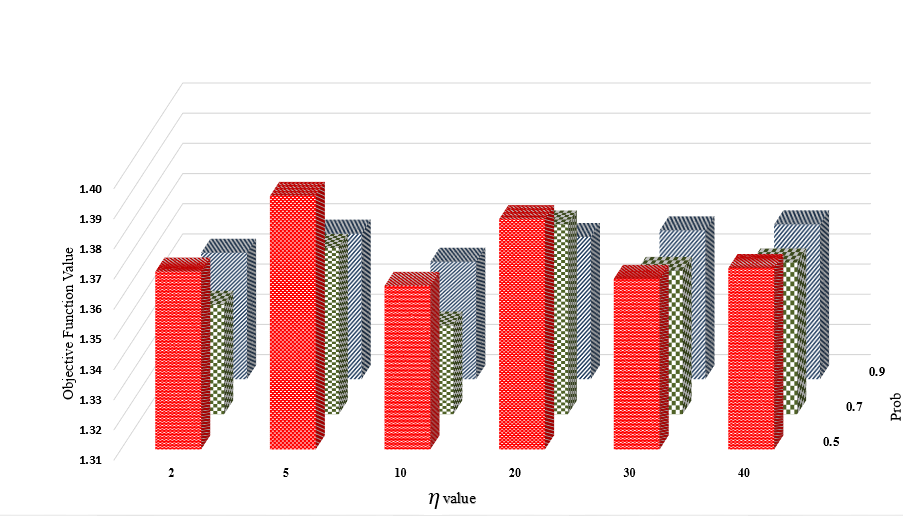}
		\caption{\textit{Barbara} image}
		\label{fig:GA_Crossover_Barbara}
	\end{subfigure}
	
	\caption{The objective function values of different \textit{prob} and $\eta$ values in the crossover operator of EnMOGA}
	\label{fig:GA_Crossover}
\end{figure}

Our crossover operator depends on two parameters, called \textit{prob} and $\eta$. To study the sensitivity of \textit{prob} and $\eta$, 18 combinations of \textit{prob} and $\eta$ are assessed ($prob =0.5, 0.7, 0.9$ and $\eta= 2,5,10,20,30,40$). All other parameters are fixed. Figures~\ref{fig:GA_Crossover} indicates the objective function value with different \textit{prob} and $\eta$ combinations for the \textit{Airplane} and \textit{Barbara} images. For the \textit{Airplane} image, we can see that the objective function value for $prob=0.5$ is higher than the other two for all $\eta$ values. There is a fluctuation in comparison between $prob=0.7$ and $prob=0.9$; meaning that for $\eta=2,5, 20$, $prob=0.7$ outperforms $prob=0.9$, while for other $\eta$ values, $prob=0.9$ provides better results than $prob=0.7$.

\begin{figure}[!htbp]
	\centering
	\begin{subfigure}[b]{0.8\linewidth}
		\includegraphics[width=\linewidth] {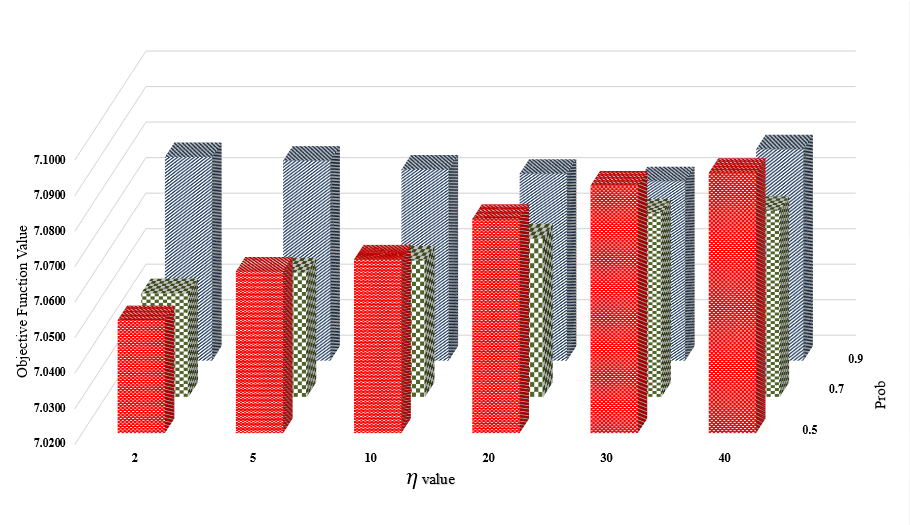}
		\caption{\textit{Airplane} image}
		\label{fig:NSGA2_Crossover_Airplane}
	\end{subfigure} 
	\begin{subfigure}[b]{0.8\linewidth}
		\includegraphics[width=\linewidth] {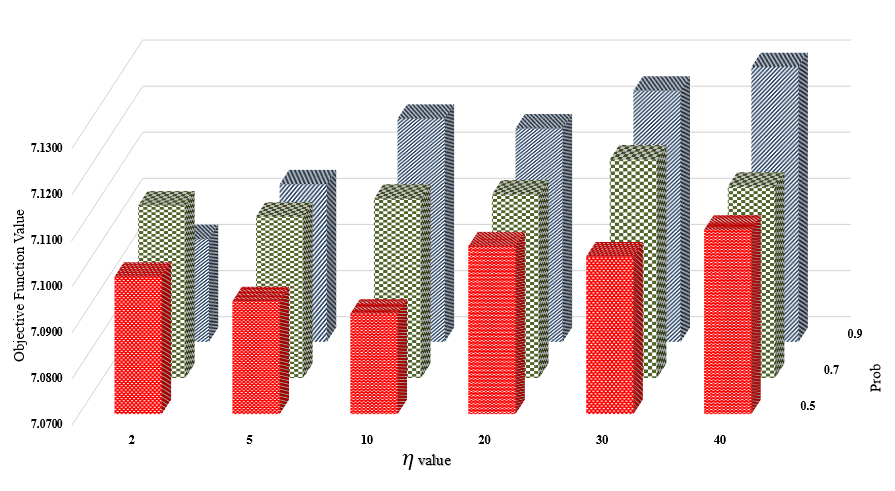}
		\caption{\textit{Barbara} image}
		\label{fig:NSGA2_Crossover_Barbara}
	\end{subfigure}
	
	\caption{The HV values of different \textit{prob} and $\eta$ values in the crossover operator of EnNSGAII}
	\label{fig:NSGA2_Crossover}
\end{figure}

This experiment is also conducted for the EnNSGAII algorithm. The results are given in Figure~\ref{fig:NSGA2_Crossover}. From Figure~\ref{fig:NSGA2_Crossover}~\subref{fig:NSGA2_Crossover_Airplane}, it is clear that $prob=0.9$ provides better results with more stability in all cases. By increasing the $\eta$ values for $prob=0.5$ and $0.7$, the HV values also are improved. The similar results can be seen in Figure~\ref{fig:NSGA2_Crossover}~\subref{fig:NSGA2_Crossover_Barbara} for the \textit{Barbara} image. In most cases, the $prob=0.9$ outperforms other $prob$ values, followed by $prob=0.7$.    

\subsubsection{Sensitivity to \textit{prob} and $\eta$ in the Mutation Operator}
\begin{figure}[!htbp]
	\centering
	\begin{subfigure}[b]{0.8\linewidth}
		\includegraphics[width=\linewidth] {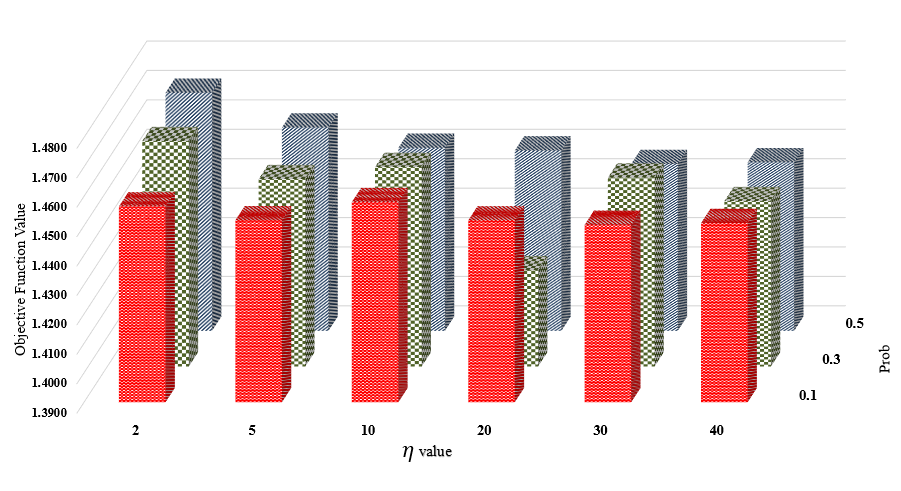}
		\caption{\textit{Airplane} image}
		\label{fig:GA_Mutation_Airplane}
	\end{subfigure} 
	\begin{subfigure}[b]{0.8\linewidth}
		\includegraphics[width=\linewidth] {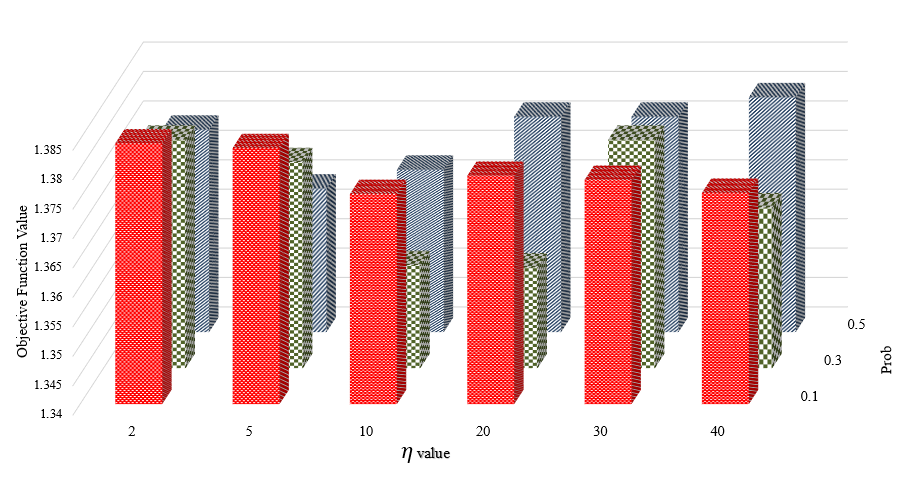}
		\caption{\textit{Barbara} image}
		\label{fig:GA_Mutation_barbara}
	\end{subfigure}
	
	\caption{The objective function values of \textit{prob} and $\eta$ in the mutation operator of EnMOGA}
	\label{fig:GA_Mutation}
\end{figure}

There are also two parameters, \textit{prob} and $\eta$, in the mutation operator. We investigated the effect of 18 combinations ($prob =0.1, 0.3, 0.5$ and $\eta= 2,5,10,20,30,40$). The results of EnMOGA are given in Figure~\ref{fig:GA_Mutation}. For the $Airplane$ image and $prob=0.5$, $\eta$ values have a downward trend; in other words, by increasing the $\eta$ values, performance is also improved. For $\eta = 0.3$, there is a fluctuation, while for $\eta = 0.1$, the results are more stable. From Figure~\ref{fig:GA_Mutation}~\subref{fig:GA_Mutation_barbara}, we can observe that the EnMOGA is sensitive to these parameters. In particular, $\eta=20$ and $prob=0.3$ provided the best results, whereas the worst results are achieved by $\eta=40$ and $prob=0.5$. 

\begin{figure}[!htbp]
	\centering
	\begin{subfigure}[b]{0.8\linewidth}
		\includegraphics[width=\linewidth] {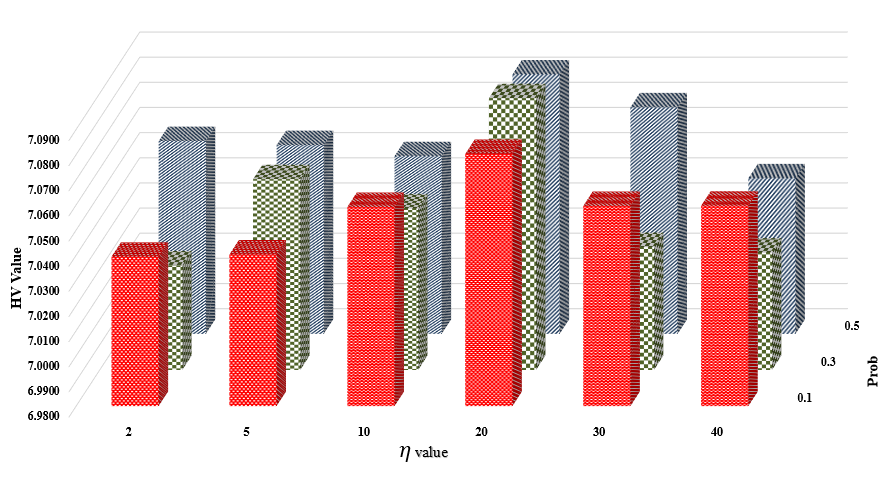}
		\caption{\textit{Airplane} image}
		\label{fig:NSGA2_Mutation_Airplane}
	\end{subfigure} 
	\begin{subfigure}[b]{0.8\linewidth}
		\includegraphics[width=\linewidth] {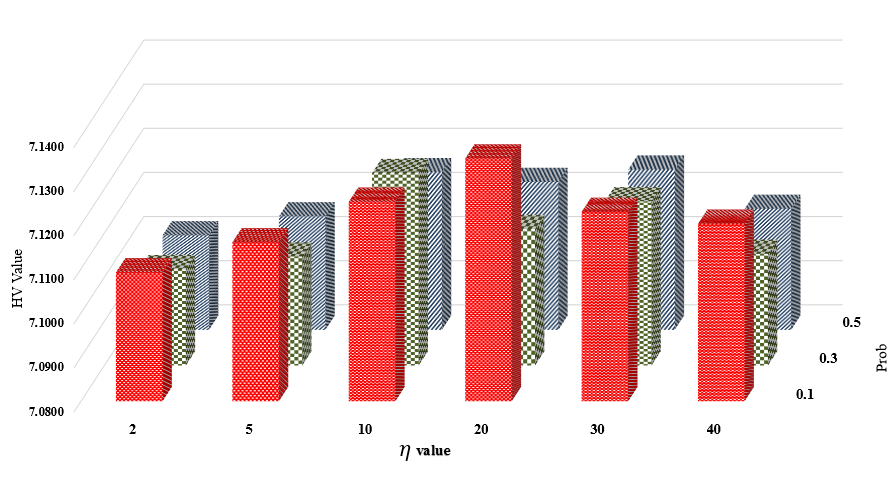}
		\caption{\textit{Barbara} image}
		\label{fig:NSGA2_Mutation_barbara}
	\end{subfigure}
	
	\caption{The HV values of \textit{prob} and $\eta$ in the mutation operator of EnNSGAII}
	\label{fig:NSGA2_Mutation}
\end{figure}

Similar results for EnNSGAII in Figure~\ref{fig:NSGA2_Mutation} indicate that, again, these parameters can affect the performance. Figure~\ref{fig:NSGA2_Mutation}~\subref{fig:NSGA2_Mutation_Airplane} investigates that $\eta=20$ can provide the highest HV values for most cases. Also, there is an upward trend from $eta=2$ to $eta=20$, while a downward trend can be seen from $eta=20$ to $eta=40$. The similar trends can also be observed in Figure~\ref{fig:NSGA2_Mutation}~\subref{fig:NSGA2_Mutation_barbara} for \textit{Barbara} image. 

\section{Conclusions}
\label{sec:Concl}
This paper proposes an energy-aware multi-objective strategy to find the best values of quantisation tables in JPEG image compression. To this end, first, we investigated whether there is a high correlation between two main properties of images, image quality and file size, and energy consumption. As a result, these two can be considered a proxy for energy consumption. Then, we defined two conflicting objective functions, including image quality and file size, while a vector-based representation was used as the candidate solution. In the next step, we embedded the proposed strategy into seven metaheuristic algorithms. Five of them are among scalarisation methods, including energy-aware multi-objective genetic algorithm (EnMOGA), energy-aware multi-objective particle swarm optimisation (EnMOPSO), energy-aware multi-objective differential evolution (EnMODE), energy-aware multi-objective evolutionary strategy (EnMOES), and energy-aware multi-objective pattern search (EnMOPS), while two others are selected among Pareto-based approaches, including energy-aware non-dominated sorting genetic algorithm (EnNSGA-II) and energy-aware reference-based NSGA-II (EnNSGA-III). Our extensive results indicated that all algorithms could outperform the baseline. In particular, EnMOGA, EnMOPS, and EnNSGA-II offered better results.

Despite the effectiveness of the proposed strategy, this work can be extended in the future with the following hints.
\begin{enumerate}
	\item This paper employed some well-established metaheuristic algorithms for the embedding process, while it can be improved by embedding the strategy into more recent algorithms such as L-SHADE~\cite{LSHADE01}.
	\item This paper ignored decomposition-based approaches (DBA) for finding the conflicting objectives, while the literature shows that DBAs have an excellent capability for multiobjective optimisation. Therefore, DBAs can be used for this problem in the future. 
	\item The goal of this paper was to find the optimal points for quantisation tables. It will likely to provide better results by adding other parameters to the current representation, such as the quality factor.
	\item This paper employed the default parameter of algorithms for the first stage of the embedding process, and we have not focused on the parameter settings for all algorithms. The optimal parameters also can be archived by a self-adaptation approach.
	\item This paper only employed two objective functions, while this research can be extended to the many-objective optimisation problem in the quantisation table generation. 
	
\end{enumerate}

\section{Acknowledgement}
This work was financed by FEDER (Fundo Europeu de Desenvolvimento Regional), from the European Union through CENTRO 2020 (Programa Operacional Regional do Centro), under project CENTRO-01-0247-FEDER-047256 – GreenStamp: Mobile Energy Efficiency Services.

This work was supported by NOVA LINCS (UIDB/04516/2020) with the financial support of FCT-Fundação para a Ciência e a Tecnologia, through national funds.

\bibliography{jalal}

\begin{thebibliography}{10}

\bibitem{DCT_Journal}
Nasir Ahmed, T\_ Natarajan, and Kamisetty~R Rao.
\newblock Discrete cosine transform.
\newblock {\em IEEE transactions on Computers}, 100(1):90--93, 1974.

\bibitem{DCT_original}
A~Andreadis, G~Benelli, A~Garzelli, and S~Susini.
\newblock A dct-based adaptive compression algorithm customized for radar
  imagery.
\newblock In {\em IGARSS'97. 1997 IEEE International Geoscience and Remote
  Sensing Symposium Proceedings. Remote Sensing-A Scientific Vision for
  Sustainable Development}, volume~4, pages 1993--1995. IEEE, 1997.

\bibitem{plot_digitiser}
{Automeris LLC}.
\newblock Webplotdigitizer.

\bibitem{JPEG_GA3}
Vinoth~Kumar Balasubramanian and Karpagam Manavalan.
\newblock Knowledge-based genetic algorithm approach to quantization table
  generation for the jpeg baseline algorithm.
\newblock {\em Turkish Journal of Electrical Engineering and Computer
  Sciences}, 24(3):1615--1635, 2016.

\bibitem{hyper_volume2}
Nicola Beume, Carlos~M Fonseca, Manuel Lopez-Ibanez, Luis Paquete, and Jan
  Vahrenhold.
\newblock On the complexity of computing the hypervolume indicator.
\newblock {\em IEEE Transactions on Evolutionary Computation},
  13(5):1075--1082, 2009.

\bibitem{pymoo}
J.~{Blank} and K.~{Deb}.
\newblock pymoo: Multi-objective optimization in python.
\newblock {\em IEEE Access}, 8:89497--89509, 2020.

\bibitem{JPEG01}
Jinyoung Choi and Bohyung Han.
\newblock Task-aware quantization network for jpeg image compression.
\newblock In {\em European Conference on Computer Vision}, pages 309--324.
  Springer, 2020.

\bibitem{IGD}
Carlos~A Coello~Coello and Margarita Reyes~Sierra.
\newblock A study of the parallelization of a coevolutionary multi-objective
  evolutionary algorithm.
\newblock In {\em Mexican international conference on artificial intelligence},
  pages 688--697. Springer, 2004.

\bibitem{JPEG_GA}
Leonardo~Faria Costa and Ant{\^o}nio Cl{\'a}udio~Paschoarelli Veiga.
\newblock Identification of the best quantization table using genetic
  algorithms.
\newblock In {\em PACRIM. 2005 IEEE Pacific Rim Conference on Communications,
  Computers and signal Processing, 2005.}, pages 570--573. IEEE, 2005.

\bibitem{Energy_APP_05}
Luis Cruz and Rui Abreu.
\newblock Catalog of energy patterns for mobile applications.
\newblock {\em Empirical Software Engineering}, 24(4):2209--2235, 2019.

\bibitem{ref_points}
Indraneel Das and John~E Dennis.
\newblock Normal-boundary intersection: A new method for generating the pareto
  surface in nonlinear multicriteria optimization problems.
\newblock {\em SIAM journal on optimization}, 8(3):631--657, 1998.

\bibitem{NSGA3}
Kalyanmoy Deb and Himanshu Jain.
\newblock An evolutionary many-objective optimization algorithm using
  reference-point-based nondominated sorting approach, part i: solving problems
  with box constraints.
\newblock {\em IEEE transactions on evolutionary computation}, 18(4):577--601,
  2013.

\bibitem{NSGA2}
Kalyanmoy Deb, Amrit Pratap, Sameer Agarwal, and TAMT Meyarivan.
\newblock A fast and elitist multiobjective genetic algorithm: Nsga-ii.
\newblock {\em IEEE transactions on evolutionary computation}, 6(2):182--197,
  2002.

\bibitem{SBX_PM}
Kalyanmoy Deb, Karthik Sindhya, and Tatsuya Okabe.
\newblock Self-adaptive simulated binary crossover for real-parameter
  optimization.
\newblock In {\em Proceedings of the 9th annual conference on genetic and
  evolutionary computation}, pages 1187--1194, 2007.

\bibitem{tutorial_statistical}
Joaqu{\'\i}n Derrac, Salvador Garc{\'\i}a, Daniel Molina, and Francisco
  Herrera.
\newblock A practical tutorial on the use of nonparametric statistical tests as
  a methodology for comparing evolutionary and swarm intelligence algorithms.
\newblock {\em Swarm and Evolutionary Computation}, 1(1):3--18, 2011.

\bibitem{Petra}
Dario Di~Nucci, Fabio Palomba, Antonio Prota, Annibale Panichella, Andy
  Zaidman, and Andrea De~Lucia.
\newblock Petra: a software-based tool for estimating the energy profile of
  android applications.
\newblock In {\em 2017 IEEE/ACM 39th International Conference on Software
  Engineering Companion (ICSE-C)}, pages 3--6. IEEE, 2017.

\bibitem{DE_Dither}
Agoston~E Eiben, James~E Smith, et~al.
\newblock {\em Introduction to evolutionary computing}, volume~53.
\newblock Springer, 2003.

\bibitem{JPEG_PSO_SA}
Pedro Henrique~Guimar{\~a}es Ferreira, Osmar Luiz~Ferreira de~Carvalho, and
  Eduardo Peixoto.
\newblock Nature inspired jpeg quantization optimization.

\bibitem{Annex_Jpeg}
O~Ferrer-Roca, RJ~Rodriguez, and A~Sousa~Pereira.
\newblock Annex x: Image formats.
\newblock In {\em Handbook of Telemedicine}, pages 252--261. IOS Press, 1998.

\bibitem{MicroEvolution02}
Viveros-Jimenez Francisco, Mezura-Montes Efren, and Gelbukh er.
\newblock Empirical analysis of a micro-evolutionary algorithm for numerical
  optimization.
\newblock {\em International Journal of Physical Sciences}, 7(8):1235--1258,
  2012.

\bibitem{android}
{Google, Jetbrains}.
\newblock Android studio.

\bibitem{Survey_MOO}
Nyoman Gunantara.
\newblock A review of multi-objective optimization: Methods and its
  applications.
\newblock {\em Cogent Engineering}, 5(1):1502242, 2018.

\bibitem{Energy_APP_09}
Geoffrey Hecht, Naouel Moha, and Romain Rouvoy.
\newblock An empirical study of the performance impacts of android code smells.
\newblock In {\em Proceedings of the international conference on mobile
  software engineering and systems}, pages 59--69, 2016.

\bibitem{PS_main_paper}
Robert Hooke and Terry~A Jeeves.
\newblock ``direct search''solution of numerical and statistical problems.
\newblock {\em Journal of the ACM (JACM)}, 8(2):212--229, 1961.

\bibitem{JPEG_SA01}
Chen-Hsiu Huang and Ja-Ling Wu.
\newblock Jqf: Optimal jpeg quantization table fusion by simulated annealing on
  texture images and predicting textures.
\newblock {\em arXiv preprint arXiv:2008.05672}, 2020.

\bibitem{PSO_Main_Paper02}
James Kennedy and Russell Eberhart.
\newblock Particle swarm optimization ({PSO}).
\newblock In {\em IEEE International Conference on Neural Networks}, pages
  1942--1948, 1995.

\bibitem{APP_01}
Hammad Khalid, Emad Shihab, Meiyappan Nagappan, and Ahmed~E Hassan.
\newblock What do mobile app users complain about?
\newblock {\em IEEE software}, 32(3):70--77, 2014.

\bibitem{hyper_volume}
Joshua Knowles and David Corne.
\newblock Properties of an adaptive archiving algorithm for storing
  nondominated vectors.
\newblock {\em IEEE Transactions on Evolutionary Computation}, 7(2):100--116,
  2003.

\bibitem{JPEG_GA2}
Mario Konrad, Herbert Stogner, and Andreas Uhl.
\newblock Evolutionary optimization of jpeg quantization tables for compressing
  iris polar images in iris recognition systems.
\newblock In {\em 2009 Proceedings of 6th International Symposium on Image and
  Signal Processing and Analysis}, pages 534--539. IEEE, 2009.

\bibitem{JPEG_DE02}
B~Vinoth Kumar and GR~Karpagam.
\newblock Knowledge-based differential evolution approach to quantisation table
  generation for the jpeg baseline algorithm.
\newblock {\em International Journal of Advanced Intelligence Paradigms},
  8(1):20--41, 2016.

\bibitem{Time_Jpeg}
B~Vinoth Kumar and GR~Karpagam.
\newblock Reduction of computation time in differential evolution-based
  quantisation table optimisation for the jpeg baseline algorithm.
\newblock {\em International Journal of Computational Systems Engineering},
  4(1):58--65, 2018.

\bibitem{JPEG_DE01}
Balasubramanian~Vinoth Kumar and Manavalan Karpagam.
\newblock Differential evolution versus genetic algorithm in optimising the
  quantisation table for jpeg baseline algorithm.
\newblock {\em International Journal of Advanced Intelligence Paradigms},
  7(2):111--135, 2015.

\bibitem{Energy_APP_01}
Ding Li and William~GJ Halfond.
\newblock An investigation into energy-saving programming practices for android
  smartphone app development.
\newblock In {\em Proceedings of the 3rd International Workshop on Green and
  Sustainable Software}, pages 46--53, 2014.

\bibitem{Earmo}
Rodrigo Morales, Rub{\'e}n Saborido, Foutse Khomh, Francisco Chicano, and
  Giuliano Antoniol.
\newblock Earmo: An energy-aware refactoring approach for mobile apps.
\newblock {\em IEEE Transactions on Software Engineering}, 44(12):1176--1206,
  2017.

\bibitem{Energy_APP_10}
Sona Mundody and K~Sudarshan.
\newblock Evaluating the impact of android best practices on energy
  consumption.
\newblock In {\em IJCA Proceedings on International Conference on Information
  and Communication Technologies}, volume~8, pages 1--4, 2014.

\bibitem{MicroEvolution01}
Mauricio Olguin-Carbajal, Enrique Alba, and Javier Arellano-Verdejo.
\newblock Micro-differential evolution with local search for high dimensional
  problems.
\newblock In {\em 2013 IEEE Congress on Evolutionary Computation}, pages
  48--54, 2013.

\bibitem{color_quantisation_SFLA}
Mar{\'\i}a-Luisa P{\'e}rez-Delgado.
\newblock Color image quantization using the shuffled-frog leaping algorithm.
\newblock {\em Engineering Applications of Artificial Intelligence},
  79:142--158, 2019.

\bibitem{JPEG_FA_TLBO}
D~Preethi and D~Loganathan.
\newblock Quantization table selection using firefly with teaching and learning
  based optimization algorithm for image compression.
\newblock In {\em Handbook of Multimedia Information Security: Techniques and
  Applications}, pages 473--499. Springer, 2019.

\bibitem{JPEG_Zigzag}
Kamisetty~Ramamohan Rao and Jae~Jeong Hwang.
\newblock {\em Techniques and standards for image, video, and audio coding}.
\newblock Prentice-Hall, Inc., 1996.

\bibitem{PSO_Main_Paper}
Yuhui Shi and Russell Eberhart.
\newblock A modified particle swarm optimizer.
\newblock In {\em IEEE International Conference on Evolutionary Computation},
  pages 69--73, 1998.

\bibitem{DE_Original}
Rainer Storn and Kenneth Price.
\newblock Differential evolution--a simple and efficient heuristic for global
  optimization over continuous spaces.
\newblock {\em Journal of Global Optimization}, 11(4):341--359, 1997.

\bibitem{LSHADE01}
Ryoji Tanabe and Alex~S Fukunaga.
\newblock Improving the search performance of shade using linear population
  size reduction.
\newblock In {\em IEEE Congress on Evolutionary Computation}, pages 1658--1665.
  IEEE, 2014.

\bibitem{JPEG_firework}
Eva Tuba, Milan Tuba, Dana Simian, and Raka Jovanovic.
\newblock Jpeg quantization table optimization by guided fireworks algorithm.
\newblock In {\em International Workshop on Combinatorial Image Analysis},
  pages 294--307. Springer, 2017.

\bibitem{JPEG_FA01}
Milan Tuba and Nebojsa Bacanin.
\newblock Jpeg quantization tables selection by the firefly algorithm.
\newblock In {\em 2014 International Conference on Multimedia Computing and
  Systems (ICMCS)}, pages 153--158. IEEE, 2014.

\bibitem{GD}
David~Allen Van~Veldhuizen.
\newblock {\em Multiobjective evolutionary algorithms: classifications,
  analyses, and new innovations}.
\newblock Air Force Institute of Technology, 1999.

\bibitem{JPEG_DE03}
B~Vinoth~Kumar and GR~Karpagam.
\newblock A smart algorithm for quantization table optimization: A case study
  in jpeg compression.
\newblock In {\em Smart Techniques for a Smarter Planet}, pages 257--280.
  Springer, 2019.

\bibitem{GA_Main_Ref}
Darrell Whitley.
\newblock A genetic algorithm tutorial.
\newblock {\em Statistics and Computing}, 4(2):65--85, 1994.

\bibitem{ES_main_paper}
Xin Yao.
\newblock Global optimisation by evolutionary algorithms.
\newblock In {\em Proceedings of IEEE International Symposium on parallel
  algorithms architecture synthesis}, pages 282--291. IEEE, 1997.

\bibitem{Adaptive_PSO}
Zhi-Hui Zhan, Jun Zhang, Yun Li, and Henry Shu-Hung Chung.
\newblock Adaptive particle swarm optimization.
\newblock {\em IEEE Transactions on Systems, Man, and Cybernetics, Part B
  (Cybernetics)}, 39(6):1362--1381, 2009.

\end{thebibliography}
\bibliographystyle{plain}

\end{document}